\documentclass[12pt,leqno]{article}
\usepackage{latexsym}

\newtheorem{theorem}{Theorem}

\newtheorem{lemma}{Lemma}

\newtheorem{definition}{Definition}

\newcommand{\blackslug}{\mbox{\hskip 1pt \vrule width 4pt height 8pt 
depth 1.5pt \hskip 1pt}}
\newcommand{\QED}{\quad\blackslug\lower 8.5pt\null\par\noindent}
\newcommand{\proof}{\par\penalty-100\vskip .5 pt\noindent{\bf Proof\/: }}
\newcommand{\ru}{\rule[-0.4mm]{.1mm}{3mm}}
\newcommand{\nni}{\ru\hspace{-3.5pt}}

\newcommand{\NI}{\mbox{$\: \nni\sim$}}

\newcommand{\cC}{\mbox{${\cal C}$}}

\newcommand{\cL}{\mbox{${\cal L}$}}
\newcommand{\cM}{\mbox{${\cal M}$}}

\newcommand{\Cn}{\mbox{${\cal C}n$}}

\newcommand{\eqdef}{\stackrel{\rm def}{=}}

\newcommand{\subseteqf}{\mbox{$\subseteq_{f}$}}

\title{Nonmonotonic Logics and Semantics
\thanks{This work was partially supported 
by the Jean and Helene Alfassa fund for 
research in Artificial Intelligence and by grant 136/94-1 of the 
Israel Science Foundation on ``New Perspectives on Nonmonotonic Reasoning''.}
}
\author{Daniel Lehmann\\
Institute of Computer Science, \\Hebrew University, \\Jerusalem 91904, Israel
\\lehmann@cs.huji.ac.il
}
\date{}
\begin{document}
\maketitle
\begin{abstract}
Tarski gave a general semantics for deductive reasoning:
a formula $a$ may be deduced from a set $A$ of formulas
iff $a$ holds in all models in which each of the elements
of $A$ holds.
A more liberal semantics has been considered: 
a formula $a$ may be deduced from a set $A$ of formulas
iff $a$ holds in all of the {\em preferred} models 
in which all the elements of $A$ hold.
Shoham proposed that the notion of 
{\em preferred} models be defined by a partial ordering on the
models of the underlying language.
A more general semantics is described in this paper, based on
a set of natural properties of choice functions.
This semantics is here shown to be equivalent to a semantics
based on comparing the relative {\em importance} of sets of models,
by what amounts to a qualitative probability measure.
The consequence operations defined by the equivalent semantics
are then characterized by a weakening of Tarski's properties
in which the monotonicity requirement is replaced by three weaker
conditions.
Classical propositional connectives are characterized by natural
introduction-elimination rules in a nonmonotonic setting.
Even in the nonmonotonic setting, one obtains classical propositional 
logic, thus showing that monotonicity is not required to justify
classical propositional connectives. 
\end{abstract}
\section{Introduction}
This paper is intended for logicians.
It builds on the insights, motivations and techniques developed
by researchers in Knowledge Representation and Artificial Intelligence,
but its purpose is to present the topic of (AI-type) nonmonotonic 
deduction (or induction) to logicians. It is not claimed that the results
of this paper will prove useful to AI practice.
It uses the language of Mathematics (theorems and proofs) to study
a form of deduction that is more general than the one used in Mathematics.
A logician interested only in the (monotonic) kind of deduction used
in Mathematics should not read further.

A semantics for nonmonotonic reasoning, more general than 
Shoham's~\cite{Shoham:87}, will be presented. This semantics is closely
related to, but generalizes, concepts developed by the Social Choice community.
In this semantic framework one may define the family of preferential
operations of~\cite{KLMAI:89} in a way that does not assume a pre-existing
monotonic logic or connectives.
Connectives may then be defined and studied by introduction-elimination
rules as is done in monotonic logics.

\section{Monotonic Logics}
\label{sec:mon}
In the thirties, Tarski made a number of fundamental advances in
the study of mathematical logic: he proposed a semantics for logical
deduction
(see in particular~\cite[p. 127]{Tarski:41}):
a formula $a$ follows from a set $A$ of formulas
iff $a$ holds in all models in which all the elements
of $A$ hold.
These ideas were first expounded in~\cite{Tar:30,Tar:30a} 
(English translation in~\cite{Tar:56}, Chapters 3 and 5 respectively).
He characterized the consequence operations that may be defined
by such a semantics
as those operations that satisfy Inclusion, Idempotence and Monotonicity,
as in Theorem~\ref{the:mon}.
That theorem, however, does not seem to appear in Tarski's work,
since he deals from the start only with compact operations.
After having settled the question of what is deduction, or what
is a logic, without being tied to any specific logical calculus,
he was able to deal with the meaning of connectives, one at a time.
This section contains a sketch of some results concerning
monotonic deductive operations,
that we are interested in generalizing
to nonmonotonic operations.

Let us assume a non-empty set (language) \cL \ and a function
\mbox{$\cC : 2^{\cL} \longrightarrow 2^{\cL}$} are given. 
Nothing is assumed about the language.
Assume \cM\ is a set (of models), about which no assumption is made,
and \mbox{$\models \: \subseteq \cM \times \cL$} is a (satisfaction)
binary relation.
For any set \mbox{$A \subseteq \cL$}, we shall denote by 
\mbox{$\widehat{A}$} or by \mbox{${\rm Mod}(A)$}
the set of all models that satisfy all elements of
$A$:
\[
\widehat{A} = {\rm Mod}(A) = 
\{x \in \cM \mid x \models a , \: \forall a \in A \}.
\]
For typographical reasons we shall use both notations, sometimes even in
the same formula.
For any set of models \mbox{$X \subseteq \cM$}, we shall denote by 
\mbox{$\overline{X}$} the set of all formulas that are satisfied 
in all elements of $X$:
\[
\overline{X} = \{a \in \cL \mid x \models a , \forall x \in X \}.
\]
The following are easily proven, for any \mbox{$A , B \subseteq \cL$},
\mbox{$X , Y \subseteq \cM$}: they amount to the fact that the operations
\mbox{$X \mapsto \overline{X}$} and \mbox{$A \mapsto \widehat{A}$}
form a Galois connection.
\[
A \subseteq \overline{\widehat{A}} \ \ \ \ , \ \ \ \ X \subseteq \widehat{\overline{X}}
\]
\[
\widehat{A \cup B} = \widehat{A} \cap \widehat{B} \ \ \ , \ \ \ \overline{X \cup Y} = 
\overline{X} \cap \overline{Y}
\]
\[
A \subseteq B \Rightarrow \widehat{B} \subseteq \widehat{A} \ \ \ , \ \ \ 
X \subseteq Y \Rightarrow \overline{Y} \subseteq \overline{X}
\]
\[
A \subseteq B \Rightarrow \overline{\widehat{A}} \subseteq \overline{\widehat{B}} \ \ \ , \ \ \ 
X \subseteq Y \Rightarrow \widehat{\overline{X}} \subseteq \widehat{\overline{Y}}
\]
\[
\widehat{A} = \widehat{\overline{\widehat{A}}} \ \ \ , \ \ \ 
\overline{X} = \overline{\widehat{\overline{X}}}
\]
\begin{theorem}
\label{the:mon}
There exists a set \cM \ (of models) and a satisfaction relation
\mbox{$\models \: \subseteq \cM \times \cL$} such that
\mbox{$\cC(A) = \overline{\widehat{A}}$} iff \cC\  satisfies the following
three conditions:
\[
{\bf Inclusion} \ \ \ A \subseteq \cC(A) , 
\]
\[
{\bf Idempotence} \ \ \ \cC(\cC(A)) = \cC(A),
\]
\[
{\bf Monotonicity} \ \ \ A \subseteq B \Rightarrow \cC(A) \subseteq \cC(B). 
\]
\end{theorem}
\proof
The {\em only if} part is very easy to prove, using the Galois connection
properties of the transformations 
\mbox{$X \mapsto \overline{X}$} and 
\mbox{$A \mapsto \widehat{A}$}.

The {\em if} part takes \cM \ to be the set of all {\em theories}, the set
of all sets \mbox{$T \subseteq \cL$} such that \mbox{$T = \cC(T)$}.
Define, then, \mbox{$T \models a$} iff \mbox{$a \in T$}.
By the definition of $\models$, \mbox{$\overline{X} = \cap_{T \in X} T$} and
\mbox{$\widehat{A}$} is the set of theories $T$ that include $A$.
\mbox{$\overline{\widehat{A}}$} is therefore the intersection of all
theories that include $A$.
Since, by Idempotence, $\cC(A)$ is a theory, and since it includes
$A$ by Inclusion, \mbox{$\overline{\widehat{A}} \subseteq \cC(A)$}.
By Monotonicity \mbox{$\cC(A) \subseteq \cC(T) = T$} for any theory
$T$ that includes $A$, and therefore $\cC(A)$ is a subset of the intersection
of all such theories $T$, i.e.,
\mbox{$\cC(A) \subseteq \overline{\widehat{A}}$}.
\QED 
As customary in the literature, \mbox{$\cC(A , B)$}
will denote \mbox{$\cC(A \cup B)$},
\mbox{$\cC(a)$} denotes \mbox{$\cC(\{a\})$} and
\mbox{$\cC(A , a)$} denotes \mbox{$\cC(A \cup \{a\})$}.
A number of important results about the operations
that satisfy Inclusion, Idempotence and Monotonicity have been proven.
Let us mention three of them, in order to consider their generalization
to nonmonotonic operations.
The first one is that an intersection of theories is a theory.
If, for any \mbox{$i \in I$}, \mbox{$A_{i} = \cC(A_{i})$},
then \mbox{$\bigcap_{i \in I} A_{i} = \cC(\bigcap_{i \in I} A_{i})$}.
The second one is that given a family $\cC_{i}$, 
\mbox{$i \in I$}, of operations that satisfy 
Inclusion, Idempotence and Monotonicity, 
their intersection, defined by,
\mbox{$(\bigcap_{i \in I} \cC)(A) \eqdef \bigcap_{i \in I} (\cC(A))$}
also satisfies 
Inclusion, Idempotence and Monotonicity.
Another result, worth noticing, is that, 
if \cC\  satisfies Inclusion, Idempotence
and Monotonicity, and if \mbox{$B \subseteq \cL$},
then the operation $\cC'$ defined by
\mbox{$\cC'(A) = \cC(A , B)$}, i.e., the operation
that follows from the acceptance of $B$ once and for all 
satisfies Inclusion, Idempotence and Monotonicity.
A last, for us, important result of Tarski that will be generalized
in Section~\ref{subsec:conn} is that the propositional
connectives may be characterized elegantly and one at a time by
Introduction-Elimination rules.
\section{Plan of this work}
\label{sec:plan}
This paper proposes a new family of operations, introduced 
in Section~\ref{sec:nonmonded}.
This family is defined by five properties, two of them
introduced here for the first time.
These properties replace Tarski's condition of Monotonicity by three
weaker properties.
The main purpose of this paper is to show the importance of this family,
and why it crops up naturally in different contexts.
To this effect, we, first, in Section~\ref{sec:simple},
describe two very different semantics, or {\em ontologies},
for nonmonotonic reasoning.
The equivalence of these semantics is proved,
under a simplifying assumption.
This equivalence lends weight
to the claim that both semantics are natural and important.
The first one is based on choice functions, enjoying properties that
have been studied by researchers 
in Social Choice (Social Preferences, Rational Choice or Revealed
Preference Theory  may have been names more to the point)
~\cite{Chernoff:54,Arrow:59,Sen:70,AizerMalish:81} already almost
half a century ago.
The
link between their preoccupations and ours certainly needs further study.
The second one is based on a qualitative notion of a measure.
Its origins may be traced to 
Ben-David and Ben-Eliyahu's~\cite{BenDavidElia:00}.
The formalism of Friedman and 
Halpern's~\cite{FrHal:PlausAAAI,FrHal:PlausJACM} is the one used here.

Then, in Section~\ref{sec:choice}, 
the simplifying logical assumption is removed and 
the properties we expect of choice functions in the more general
framework are discussed.
One of the central ideas necessary to deal with the general
case {\em definability preservation} has been put in evidence by
Schlechta's~\cite{Schle:91}.
The semantics proposed are a natural generalization of Tarski's semantics.
The interest and importance of the family of operations described in 
Section~\ref{sec:nonmonded}
are put in evidence in Section~\ref{sec:mainchar}, where it is shown
that they characterize exactly the operations defined by the semantics
of Section~\ref{sec:choice}.
In Section~\ref{sec:Properties} we discuss further properties of the operations
of the family defined in Section~\ref{sec:nonmonded} and, in particular,
the characterization of propositional connectives by Introduction-Elimination
rules. It is shown that the logics of the (semantically) classical
connectives is classical propositional calculus.
In Section~\ref{sec:KLM} it is shown that, if the language \cL\ is
a propositional calculus, the finitary consequence
relations defined by our family of operations are exactly the 
preferential consequence relations of~\cite{KLMAI:89}.
In Section~\ref{sec:rat} a more restricted family of operations,
defined by equivalent semantic restrictions concerning choice functions
on one hand and qualitative measures on the other hand,
is characterized by an additional requirement on the nonmonotonic operations.
Properties of this family are briefly discussed.
Section~\ref{sec:conc} is a conclusion. 
\section{Nonmonotonic Deduction Operations}
\label{sec:nonmonded}
In this section we shall present five properties of an operation
\mbox{$\cC : 2^{\cL} \longrightarrow 2^{\cL}$}, discuss them
and their relation to properties of monotonic operations.
We shall argue that they should be satisfied by inference operations.
In the next sections, we shall define suitable semantics for them
and prove a representation theorem.

Our first two properties are uncontroversial, at least for mathematical
logicians. They may not be satisfied by most deductive agents
with bounded resources, though.
Inclusion and Idempotence are two of the conditions characterizing
monotonic deduction and seem most natural also in the context
of nonmonotonic deduction.
\[
{\bf Inclusion} \ \ \ \forall A \subseteq \cL\ 
A \subseteq \cC(A) , 
\]
\[
{\bf Idempotence} \ \ \ \forall A \subseteq \cL\ 
\cC(\cC(A)) = \cC(A).
\]
The first one, Inclusion, requires that all assumptions be
part of the conclusions.
The second one, Idempotence, expresses the requirement
that the strength or the validity of a conclusion be unaffected
by the length of the chain of arguments leading to its acceptance.
We require the operation \cC\  squeeze the {\em fruit}, i.e., the
assumptions to the end, i.e., until no more conclusions can
be obtained.

The next three properties are properties of restricted monotonicity.
They are implied by Monotonicity, and express that, in certain situations,
Monotonicity is required.
\[
{\bf Cautious \ Monotonicity} \ \ \ \forall A , B \subseteq \cL\ 
A \subseteq B \subseteq \cC(A) 
\Rightarrow \cC(A) \subseteq \cC(B) 
\]
Cautious Monotonicity is a restricted form of Monotonicity:
any monotonic operation is cautiously monotonic.
This property was first introduced by D. Gabbay~\cite{Gabbay:85},
in its finitary form and by D. Makinson~\cite{Mak:89,Mak:Handbook}
in its infinitary form.
It requires that one does not retract previous conclusions when
one learns that a previous conclusion is indeed true.
It seems to have been accepted as reasonable by all researchers in the field.
A discussion of its appeal may be found in~\cite{KLMAI:89}.

The next two properties are described here for the first time,
but they are closely related to the property previously discussed under
the name of Deductivity or Infinite Conditionalization
in~\cite{FL:JELIA90,FL:IGPL,Freund:supra93,FL:Studia,Mak:Handbook,KalLeh:94}.
The first one, termed Conditional Monotonicity, expresses the requirement that
\cC\  behave monotonically if one replaces, in the assumptions, some of
the assumptions by their consequences.
It asserts that non-monotonicity cannot be caused by the deduction process
itself. It is only the addition of new assumptions unrelated to the old
ones or assumptions that are less than the full sets of conclusions
that can lead to non-monotonicity. 
\[
{\bf Conditional \ Monotonicity} \ \ \ \ \forall A , B \subseteq \cL, \ 
\cC(A , B) \subseteq \cC(\cC(A) , B)
\]
This is indeed a monotonicity requirement, if \cC\  satisfies Inclusion,
since \mbox{$A \cup B \subseteq \cC(A) \cup B$}.
But this is a very restricted form of Monotonicity:
we do not allow adding arbitrary, unrelated, assumptions, only replacing
part of the assumptions by their consequences.
When doing so, we do require that one does not lose conclusions,
but one may add conclusions.
It is worth noticing that Conditional Monotonicity is an intrinsically
infinitary condition: even if $A$ and $B$ are finite, $\cC(A)$ that
appears on the right hand side is typically infinite.
If \cC\ is a Tarski deductive operation, i.e., 
satisfies Inclusion, Idempotence and Monotonicity, then, 
\[
\cC(A , B) \subseteq \cC(\cC(A) , B) \subseteq
\cC(\cC(A) , \cC(B)) \subseteq \cC(\cC(A , B)) \subseteq 
\cC(A , B)
\]
and we have:
\mbox{$\cC(A , B) = \cC(\cC(A) , B)$}.
In the system presented here, \mbox{$\cC(A , B)$} may be a strict
subset of \mbox{$\cC(\cC(A) , B)$}.
Conditional Monotonicity seems related with Cautious Monotonicity, 
but it is not.
In Cautious Monotonicity, we allow the addition of a subset of $\cC(A)$,
whereas Conditional Monotonicity requires the addition of $\cC(A)$ in its entirety.
In Conditional Monotonicity, the addition may be done in the presence of some
other assumptions ($B$), whereas in Cautious Monotonicity $B$ must be
empty.
The intuitive justification for Conditional Monotonicity will be given now,
it explains the term Conditional Monotonicity:
if \mbox{$c \in \cC(A , B)$}, then, in $\cC(A)$, there should be
something (some conditional statements)
to the effect that: {\em if} $B$ {\em then} $c$.
But, together with $B$, this should imply $c$ and therefore $c$
should be in \mbox{$\cC(\cC(A) , B)$}.

Our last property, termed Threshold Monotonicity, requires that
\cC\  behave monotonically in all cases in which the deduction process
has already been applied to part of the assumptions,
i.e., above the threshold of some $\cC(A)$.
If $\cC(A)$ is part of the assumptions, then \cC\  behaves monotonically.
\[
{\bf Threshold \ Monotonicity} \ \ \ \  \cC(A) \subseteq B \subseteq C 
\: \Rightarrow \:
\cC(B) \subseteq \cC(C)
\]
The intuitive reason for this requirement is similar to the one for
Conditional Monotonicity.
We expect that, if \mbox{$\cC(A) \subseteq B$} and
\mbox{$c \in \cC(B)$}, formulas saying that:
{\em if} $B$ {\em then} $c$, are already in $\cC(A)$.
The set $C$, then, contains both {\em if} $B$ {\em then} $c$ and 
$B$, therefore \mbox{$\cC(C)$} should contain $c$.

Before we move to our main representation result, let us draw one
important consequence of the properties above.
\begin{lemma}[Cumulativity, Makinson~\cite{Mak:Handbook}]
\label{le:cum}
If \cC\  satisfies Idempotence and Cautious Monotonicity,
then it satisfies
\[
{\bf Cumulativity} \ \ \ 
A \subseteq B \subseteq \cC(A) \Rightarrow \cC(B) = \cC(A).
\]
\end{lemma}
The importance of Cumulativity has been stressed early on
by Makinson~\cite{Mak:89}.
\proof
By Cautious Monotonicity, \mbox{$\cC(A) \subseteq \cC(B)$}.
Therefore we have \mbox{$B \subseteq$} \mbox{$\cC(A) \subseteq$}
\mbox{$\cC(B)$}.
By Cautious Monotonicity again, we have
\mbox{$\cC(B) \subseteq$} \mbox{$\cC(\cC(A))$}.
By Idempotence, then
\mbox{$\cC(B) \subseteq \cC(A)$}.
\QED
\section{Two semantics in a simplified framework}
\label{sec:simple}
We shall now describe two natural different semantics for nonmonotonic
operations: one based on choice functions and one based on qualitative
probability measures. 
We shall show their equivalence, under simplifying assumptions to be described
now.
The exact fit between the semantics based on choice functions and the formal
properties of nonmonotonic deduction operations described in 
Section~\ref{sec:nonmonded} will be proved in Section~\ref{sec:mainchar}.
\subsection{A simplifying assumption}
\label{subsec:simple}
The general setting, assuming an arbitrary language \cL,
will be developed in 
Section~\ref{sec:choice}.
It requires that sets of models that may be defined by a set
of formulas be given prominence.
\begin{definition}
\label{def:definable}
A set $X$ of models is said to be definable iff one of the two
following equivalent conditions holds: 
\begin{enumerate}
\item
\mbox{$\exists A \subseteq \cL$} such that \mbox{$X = \widehat{A}$}, or
\item
\mbox{$X =  \widehat{\overline{X}}$}.
\end{enumerate}
The set of all definable subsets of $X$ will be denoted by $D_{X}$.
\end{definition} 
A set of models is definable iff it is the set of all models
satisfying some set of formulas.
In many situations in which 
a {\em finiteness} assumption is reasonable,
one may avoid the consideration of
the special role of definable sets.
For example, researchers in Social Choice typically assume the set
of outcomes is finite.
Friedman and Halpern, also, assume, at least in part of their work,
that the base set is finite.
For expository purposes, to keep definitions and justifications simple, 
we shall now make a similar assumption,
to be lifted in Section~\ref{sec:choice}.
We shall assume that every set of models is definable. 
\[
{\bf Simplifying \ Assumption \ } \forall X \subseteq \cM , \: 
X = \widehat{\overline{X}}
\] 
Any propositional calculus on a {\em finite} number of atomic
propositions satisfies our Simplifying Assumption.
\subsection{Choice functions}
\label{subsec:simplechoice}
The properties described in this section have been put in evidence
for the first time, probably, by researchers in social choice,
triggered by H. Chernoff~\cite{Chernoff:54}, H. Uzawa~\cite{Uzawa:56} and 
K. Arrow~\cite{Arrow:59}
(for an updated survey, see~\cite{Moulin:85}).
The exact nature of the link between nonmonotonic logics and the theory
of choice functions needs further research.
For the sake of those readers who are not familiar with this literature,
let us describe briefly its framework: each individual has personal preferences
over a set of possible outcomes. The society, given a subset of those,
the feasible outcomes, must come up with a the subset of those feasible
outcomes that are acceptable socially, in view of the individual preferences.
Different methods of social decision result in different functions
from sets of feasible outcomes to sets of acceptable outcomes.
Social Choice investigates the relations between those
different methods for social decision and the choice functions
they determine.

Independently, Y. Shoham, in~\cite{Shoham:87}, 
proposed a general semantics for
nonmonotonic reasoning, based on preferences among models.
The link between the properties of choice functions studied by Social
Choice researchers and Nonmonotonic Reasoning has been put in evidence
by Doyle and Wellman~\cite{DoyleWell:91}, Rott~\cite{Rott:93} 
and Lindstr\"{o}m~\cite{Lindstrom:94}. 
Lindstr\"{o}m generalizes the finitary framework considered in Revealed
Preference Theory to an infinitary framework. 
In~\cite{Schle:91,Schlech:95}, Karl Schlechta rediscovered choice functions
and their properties, in the infinitary framework.
He also considered an additional property needed in such a framework.
Schlechta's line of research is best described in~\cite{Schlechta:SPLN}
and~\cite{Schlechta:JSL}.
The only novelty of this section is the detailed argumentation
justifying the assumptions about the choice function.

The basic idea is that one can generalize Tarski's 
semantic analysis
of deduction by considering, instead of all models
of a set $A$ of formulas, only a subset of this set:
the set of {\em best} models of $A$, jumping, on the basis of $A$, 
to the conclusion that the situation at hand is one of those {\em best}
situations. 
Such a manner of drawing conclusions from categorical information
has been time and again attested by researchers in cognitive sciences.
Lakoff's~\cite{Lakoff:Women} is a good introduction.
The author's~\cite{Leh:Ster} represents a very tentative formalization.

We consider a set \cM\ 
of models, a satisfaction relation $\models$ and a choice function
\mbox{$f : 2^{\cM} \longrightarrow 2^{\cM}$} that chooses,
for a set $X$ of models, the set of {\em best}, {\em most typical},
{\em most important}, or {\em preferred} models for the set $X$.
Then we define the function \cC\  by 
\begin{equation}
\label{eq:fdef}
\cC(A) = \overline{f(\widehat{A})}. 
\end{equation}
The use of a choice function to define a (nonmonotonic) consequence
operation generalizes Shoham's semantics and represents one step
up in the abstraction ladder from Shoham's semantics. 
Shoham assumed that the choice function $f$ is defined in a particular
way: $f(X)$ is the set of all elements of $X$ that are {\em minimal}
in $X$, under some, pre-existing, order relation on $\cM$.
We prefer to deal directly with properties of the function $f$,
without assuming any order relation on $\cM$.

Let us now present natural conditions on $f$.
The monotonic case corresponds to the case that the function $f$ 
is required to be the identity function. The conditions below are
trivially satisfied by the identity function.

Let us take the following running example, that will exemplify
the fact that the properties of the choice function $f$ define,
in a sense the logic of optimization.
You are looking for an apartment in Paris.
The set of apartments on the market in Paris is the set \cM.
Your real estate agent asks for your desiderata and your financial
possibilities. You expect her to come up with a limited list
of the {\em best} apartments available in Paris.
You expect a list that is neither too small nor too large.
Notice that this example does not fit exactly the Social Choice paradigm,
interested in finding a subset of outcomes that are acceptable
in the view of contradictory preferences of the individuals,
since we assumed you alone are looking for an apartment.
But it may be fitted to the Social Choice paradigm if you consider
that you are shopping for an apartment that fits in some way the
contradictory desiderata of a family, and of the real estate agent.

The first property we expect of $f$ is a property of contraction.
\[
{\bf Contraction} \ \ \ f(X) \subseteq X
\]
Indeed, intuitively, $f$ picks the {\em preferred} models of the
set $X$ and those models are in the set $X$.
The identity function satisfies Contraction.
Contraction is assumed in the Social Choice literature without even 
mentioning it explicitly.

If you requested an apartment in Paris, you expect to get
a list of apartments in Paris, not in Neuilly or in San Francisco.

The second property expresses the fact that, if $X$ is a subset of $Y$,
it is more difficult to be one of the best of the bigger set $Y$ than to
be one of the best of the smaller set $X$. Therefore we expect any 
element of $f(Y)$ that happens to be in $X$ to be in $f(X)$.
\[
{\bf Coherence} \ \ \ X \subseteq Y \Rightarrow X \cap f(Y) \subseteq f(X)
\]
This is Sen's~\cite{Sen:70} property $\alpha$.
The identity function satisfies Coherence.
Coherence is a kind of antimonotonicity:
if \mbox{$X \subseteq Y$}, then antimonotonicity would require:
\mbox{$f(Y) \subseteq f(X)$}, whereas Coherence only requires
that this part of $f(Y)$ that is included in $X$ is included in $f(X)$.
The term Coherence seems to be an appropriate name since it expresses
the existence of some kind of coherent test by which the {\em preferred}
elements of a set are picked up: the test corresponding to a superset
must be at least as demanding as the one of a subset. 
The Coherence property
appears in Chernoff's~\cite{Chernoff:54} and has been given his name 
in~\cite{Moulin:85}. It has been endorsed by all
researchers in Social Choice.

To illustrate Coherence, 
suppose you now remember you promised your wife you would live on the 
left bank, but forgot to tell that to your agent. You tell her that
and get a new list, of apartments on the left bank.
You expect all the left bank apartments that appeared
in the first list to be included in the second list.
The second list will probably also include other apartments, that were not
part of the, say twenty, best apartments in Paris, but are part the best
apartments on the left bank.

The third and last property expresses the fact that, if $Y$ is a subset of $X$,
but large enough to include all of $f(X)$, then we do not expect
$f(Y)$ to be larger than $f(X)$.
\[
{\bf Local \ Monotonicity} \ \ \ \ f(X) \subseteq Y \subseteq X \Rightarrow
f(Y) \subseteq f(X)
\]
The term Local Monotonicity expresses that this is a property
of qualified monotonicity for $f$: \mbox{$Y \subseteq X$} implies
\mbox{$f(Y) \subseteq f(X)$}, conditional on 
\mbox{$f(X) \subseteq Y$},
somehow a local condition.
The importance of property has been put in evidence
by M. A. Aizerman~\cite{AizerMalish:81,Aizer:85}.
The identity function satisfies Local Monotonicity, because it is monotonic,
but also trivially, because the assumption implies \mbox{$X = Y$}.

Suppose all best apartments of Paris included in the
list you got from your agent happen to be on the left bank.
You certainly would not expect a larger list if you told her you
want only apartments on the left bank.

Any choice function satisfying Contraction, Coherence and Local Monotonicity
is considered acceptable in this work.
A number of other properties of choice functions, from the literature, 
will be discussed now.
Researchers in Social Choice have universally endorsed the following:
$f(X)$ is not empty if $X$ is not empty.
Some have noticed that this property is not crucial and that they could
build the theory without it. We have no reason to make this requirement:
it may well be the case that our search for a large cheap apartment in Paris
in the best quarter leaves us with an empty list.

Another property that has been widely considered is:
\[
{\bf Expansion} \ \ f(X) \cap f(Y) \subseteq f(X \cup Y)
\] 

It does not follow from Contraction, Coherence and Local Monotonicity.
We do not endorse it. An apartment, on the Ile de la Cit\'{e},
that makes the list of the
ten best apartments on the left bank and (or) the Ile de la Cit\'{e}
and also makes the list of the ten best apartments on the right bank 
and (or) the Ile de la Cit\'{e}, does not necessarily makes the list
of the ten best apartments in Paris.
One easily sees that the semantics proposed by Shoham in~\cite{Shoham:87} 
validates Contraction, Coherence, Local Monotonicity and also
Expansion: if $z$ is minimal in $X$ and minimal in $Y$ it is
minimal in the union \mbox{$X \cup Y$}.
Therefore, the semantics proposed in this paper is a strict generalization 
of Shoham's.
In fact, under our simplifying assumption, Shoham's semantics
is equivalent to considering choice functions that satisfy Contraction,
Coherence, Local Monotonicity and Expansion.

We have presented a semantic framework that generalizes Tarski's.
It involves a choice function on sets of models.
This choice function is assumed to satisfy three conditions.
We argued that those three conditions are natural, but the
ultimate test of their interest lies ahead.
The Social Choice literature mentions two main results about choice functions
that satisfy Contraction, Coherence and Local Monotonicity.
In~\cite{Plott:73}, it is shown that they are exactly the choice functions
that satisfy Contraction and 
\[
{\bf Path \ Independence} \ \ \ f(X \cup Y) = f(f(X) \cup Y). 
\]
In~\cite{AizerMalish:81}, it is shown that, if \cM\ is finite, 
they are exactly the 
{\em pseudo-rationalizable} choice functions, i.e., 
those that may be defined by a finite set of binary {\em preference}
relations $>_{i}$ on \cM\  by taking, for $f(X)$, the set of all elements of $X$
that are minimal in $X$ for at least one of the $>_{i}$'s.
None of these results will be used in this paper.
\subsection{Qualitative Measures}
\label{subsec:qualmeas}
A completely different generalization of Tarski's semantic analysis
will be reviewed now. 
Its origins may be traced to Dubois and Prade~\cite{DubPrade:IJCAI91}
and Ben-David and 
Ben-Eliyahu~\cite{BenDavidElia:00}.
Up to small technical changes, our presentation will be that of
Friedman and Halpern~\cite{FrHal:PlausAAAI,FrHal:PlausJACM}.
The connection between both approaches is described 
in~\cite{Schlechta:Filters}. Some more results concerning the link
between plausibility measures and preferential relations may be found 
in~\cite{FreundJLC:98}.

Suppose we had some way of measuring the {\em size} or the
{\em importance} of sets of models.
One is tempted to say that a formula $a$ may be deduced
from a set $A$ of formulas iff the measure of the set of all models of $A$ 
that satisfy $a$ is larger than that of the set of models of $A$ that do not
satisfy $a$.
Since, as in the case of monotonic logic, 
one would like to deduce anything from an inconsistent set
of formulas, the case that the set of models of $A$ is negligible, 
i.e., not larger than the empty set, has to be treated separately.
With the notations of Section~\ref{subsec:simplechoice}, 
one would like to assume
a binary relation $>$ on $2^{\cM}$ (\mbox{$X > Y$} iff the measure of
$X$ is larger than that of $Y$) and define the deductive operation by:
\begin{equation}
\label{eq:secdef}
a \in \cC(A) \ {\rm iff} \ {\rm either} \ 
\widehat{A} \cap \widehat{\{ a \}} > \widehat{A} - \widehat{\{ a \}}
\ {\rm or} \ \widehat{A} \not > \emptyset.
\end{equation}
This definition indeed generalizes Tarski's.
For Tarski: \mbox{$X > Y$} iff \mbox{$Y = \emptyset$} and
\mbox{$X \neq \emptyset$}.

The following list of natural properties for $>$ is similar to
Friedman and Halpern's definition of a Qualitative Plausibility
Measure.\footnote{Friedman and Halpern assume a function
\mbox{$Pl : 2^{\cM} \mapsto D$} and a reflexive, transitive
and anti-symmetric relation $\leq$ on $D$ satisfying:
\begin{itemize}
\item {\bf A1} if \mbox{$X \subseteq Y$} then \mbox{$Pl(X) \leq Pl(Y)$},
\item {\bf A2} if \mbox{$X , Y , Z$} are pairwise disjoint sets,
\mbox{$Pl(X \cup Y) \geq Pl(Z)$} and \mbox{$Pl(X \cup Z) \geq Pl(Y)$},
then \mbox{$Pl(X) \geq Pl(Y \cup Z)$},
\item {\bf A3} if \mbox{$Pl(X) = Pl(Y) = Pl(\emptyset)$}, then
\mbox{$Pl(X \cup Y) = Pl(\emptyset)$}.
\end{itemize}
}
The properties described here are in fact stronger than
theirs: the representation result holds for both sets of properties.
First, it seems reasonable to require that $>$ be a strict partial order
relation, i.e., irreflexive and transitive. This seems implied
by our description of \mbox{$X > Y$} as meaning that the {\em measure}
of $X$ is larger than that of $Y$.
Note that we do not require the relation $>$ to be total, which would 
not be reasonable since $X$ and $Y$ may have equal measure without
being equal, nor even to be {\em modular}, i.e., satisfy the property
\mbox{$X > Y$} implies that, for any $Z$, either \mbox{$X > Z$}
or \mbox{$Z > Y$}. Modularity will be assumed in Section~\ref{sec:rat}.
\begin{equation}
\label{eq:porder}
{\rm The \ relation \ } > {\rm \ is \ irreflexive \ and \ transitive.}
\end{equation}
A second very natural property of $>$ is that it should behave 
as expected with respect to set inclusion.
The reader will easily be convinced that, in our setting, the correct
formulation is the following:
\begin{equation}
\label{eq:inc}
W \supseteq X > Y \supseteq Z \: \Rightarrow \: W > Z.
\end{equation}
Our third property deals with the special character of the empty set.
The empty set is the ultimate {\em small} set and any set $X$ that
is not strictly greater than the empty set must be extremely small
and negligible.
In many cases, any nonempty set will be strictly greater than the empty set,
but we do not wish to make this a requirement.
Since the union of any family of empty sets is empty, it is reasonable
to require that
the union of a family of negligible sets be negligible.
\begin{equation}
\label{eq:small}
\forall i \in I , X_{i} \not > \emptyset \: \Rightarrow \: 
\bigcup_{i \in I} X_{i} \not > \emptyset
\end{equation}
This is the infinitary version of property (A3) of Friedman and Halpern.
The finitary version is not enough for Theorem~\ref{the:plch} to hold.
The next two properties have to deal with the qualitative character
of the relation $>$. Qualitatively greater has to be understood here
as {\em an order of magnitude} greater.
Assume \mbox{$X \cup Y > Y$}. If \mbox{$X \cup Y$} is an order of magnitude
greater than $Y$, it must be that $X$ is already greater than $Y$.
\begin{equation}
\label{eq:in}
X \cup Y > Y \: \Rightarrow \: 
X > Y
\end{equation}
Note that the definition of \cC\ in~\ref{eq:secdef}
makes use of the relation $>$ only between sets with an empty intersection.
Property~\ref{eq:in} can therefore only have an indirect influence.
The qualitative plausibility measures of Friedman and Halpern
need not satisfy Property~\ref{eq:in}.
The results presented in Section~\ref{subsec:equiv} show that
one may add this property without harm.
Friedman and Halpern consider a property (A2) that implies 
the finitary version of~\ref{eq:union}.
The finitary version of~\ref{eq:union} together with~\ref{eq:in}
imply (A2).
The next and last property is the fundamental one that makes the qualitative
character of $>$ apparent.
\begin{equation}
\label{eq:union}
\forall i \in I , \: X > Y_{i}  \: \Rightarrow \: 
X > \bigcup_{i \in I} Y_{i}
\end{equation}
A set that is greater than every one of a family of sets must
be greater than their union: pooling small sets never makes a big set.
This seems to be the essence of {\em qualitative}.

We have presented a set of properties for $>$ and have argued that
they are natural properties for a qualitative measure.
No such argument can be completely convincing.
In the next section, it will be shown that the 
properties~\ref{eq:porder}, \ref{eq:inc}, \ref{eq:small}, \ref{eq:in}
and~\ref{eq:union} for $>$ are equivalent to the conditions
of Contraction, Coherence and Local Monotonicity for $f$.
This equivalence suggests that those two sets of properties
and the nonmonotonic operations they define have a central role to
play in the study of nonmonotonic logics.
To the best of my knowledge, the qualitative measures have not been
studied by the Social Choice community.
Since, as will be seen in Section~\ref{subsec:equiv}, they are equivalent
to choice functions, it may be worthwhile to ask whether they can
help there.
The property of Expansion considered in Social Choice translates
readily into 
\mbox{$\bigcup_{i \in I} X_{i} > Z$} implies that there is some
\mbox{$i \in I$} such that \mbox{$X_{i} > Z$}.
\subsection{A Semantic Equivalence}
\label{subsec:equiv}
\begin{theorem}
\label{the:chpl}
Suppose $f$ is a choice function that satisfies Contraction, Coherence
and Local Monotonicity.
Then, the relation $>$ defined by:
\mbox{$X > Y$} iff \mbox{$f(X) \neq \emptyset$} and 
\mbox{$Y \cap f(X \cup Y) = \emptyset$} satisfies 
properties~\ref{eq:porder}, \ref{eq:inc}, \ref{eq:small}, \ref{eq:in}
and~\ref{eq:union}.
If \cC\ is defined by Equation~\ref{eq:fdef}, then it satisfies 
Equation~\ref{eq:secdef}.
\end{theorem}
Before we get to the proof, let us ponder on the translation
proposed. Is it a natural translation, i.e., does it fit the intuitive
interpretations given to $>$ and $f$?
The sets $X$ such that \mbox{$f(X) = \emptyset$} are the {\em no solution}
sets (in our running examples: no suitable apartments in $X$).
They may be assumed to be small.
If $X$ is qualitatively larger than $Y$, it is reasonable to assume that
$X$ is not so small as to have an empty image under $f$. 
In Section~\ref{subsec:qualmeas}, the intuition we developed
was that $X$ is qualitatively larger than $Y$ means that $X$ contains some
important elements $Y$ does not contain.
The important elements are those of $f(X)$ and $Y$ contains none of them.
\proof
Let us show, first, that \cC\  satisfies Equation~\ref{eq:secdef}.
Obviously \mbox{$a \in \overline{f(\widehat{A})}$} iff
\mbox{$f(\widehat{A}) \subseteq \widehat{\{ a \}}$}.
By Contraction, we have
\[
f(\widehat{A}) \subseteq \widehat{\{ a \}}
\ {\rm iff} \ 
(\widehat{A} - \widehat{\{ a \}}) \cap f(\widehat{A}) = \emptyset.
\]
The only thing left for us to check is that if
\mbox{$(\widehat{A} - \widehat{\{ a \}}) \cap f(\widehat{A}) = \emptyset$}
and \mbox{$f(\widehat{A}) \neq \emptyset$}, then
\mbox{$f(\widehat{A} \cap \widehat{\{ a \}})\neq \emptyset$},
which follows from Coherence.

Irreflexivity of $>$ follows from Contraction.
For Transitivity, assume \mbox{$f(X) \neq \emptyset$},
\mbox{$Y \cap f(X \cup Y) = \emptyset$}, \mbox{$f(Y) \neq \emptyset$}
and \mbox{$Z \cap f(Y \cup Z) = \emptyset$}.
We want to show that \mbox{$Z \cap f(X \cup Z) = \emptyset$}.
By Contraction, we know that
\mbox{$f(X \cup Y \cup Z) \subseteq$}
\mbox{$(X - Z) \cup$}
\mbox{$(Y \cup Z) \cap f(X \cup Y \cup Z)$}.
By Coherence, 
\mbox{$(Y \cup Z) \cap f(X \cup Y \cup Z) \subseteq$}
\mbox{$f(Y \cup Z)$}.
By assumption, \mbox{$f(Y \cup Z) \subseteq$} \mbox{$Y - Z$}.
We conclude that \mbox{$f(X \cup Y \cup Z) \subseteq$} \mbox{$(X \cup Y) - Z$}
and therefore \mbox{$f(X \cup Y \cup Z) \subseteq$} \mbox{$X \cup Y$}.
By Coherence again, then
\mbox{$f(X \cup Y \cup Z) \subseteq$} \mbox{$f(X \cup Y)$} and, by assumption,
\mbox{$f(X \cup Y \cup Z) \subseteq X$}.
We have 
\mbox{$f(X \cup Y \cup Z) \subseteq$} 
\mbox{$X \cup Z \subseteq$}
\mbox{$X \cup Y \cup Z$}.
By Local Monotonicity, then, we conclude that
\mbox{$f(X \cup Z) \subseteq$} \mbox{$f(X \cup Y \cup Z)$}.
Therefore \mbox{$f(X \cup Z) \subseteq$} \mbox{$(X \cup Y) - Z$}.
We conclude that \mbox{$Z \cap f(X \cup Z) = \emptyset$}.

For property~\ref{eq:inc}, assume \mbox{$X \supseteq Y$},
\mbox{$W \supseteq Z$}, \mbox{$f(Y) \neq \emptyset$} and
\mbox{$W \cap f(Y \cup W) = \emptyset$}.
Assume \mbox{$f(X) = \emptyset$}.
We have \mbox{$f(X) \subseteq$}
\mbox{$Y \subseteq X$} and,
by Local Monotonicity, \mbox{$f(Y) \subseteq f(X)$} and
\mbox{$f(X) \neq \emptyset$}.
We have shown that \mbox{$f(X) \neq \emptyset$}.
We have \mbox{$f(Y \cup W) \subseteq$} 
\mbox{$Y \subseteq$} \mbox{$Y \cup Z \subseteq$}
\mbox{$Y \cup W$}.
By Local Monotonicity, then, we have
\mbox{$f(Y \cup Z) \subseteq$} \mbox{$f(Y \cup W)$}.
By Coherence,
\[
(Y \cup Z) \cap f(X \cup Z) \subseteq f(Y \cup Z) \subseteq
f(Y \cup W) \subseteq Y - W \subseteq Y - Z.
\]
We conclude that \mbox{$Z \cap f(X \cup Z) = \emptyset$}.

For property~\ref{eq:small}, we notice that, from the definition of $>$, 
\mbox{$X > \emptyset$} iff \mbox{$f(X) \neq \emptyset$}.
We have to show that \mbox{$f(X_{i}) = \emptyset$} for any
\mbox{$i \in I$}
implies \mbox{$f(\bigcup_{i \in I} X_{i}) = \emptyset$}.
This follows from Contraction and Coherence.

For property~\ref{eq:in}, assume \mbox{$f(X \cup Y) \neq \emptyset$}
and \mbox{$Y \cap f(X \cup Y) = \emptyset$}.
The only thing left for us to prove is that we have 
\mbox{$f(X) \neq \emptyset$}. 
Notice that, by Coherence, \mbox{$X \cap f(X \cup Y) \subseteq$}
\mbox{$f(X)$}. It is enough to show that 
\mbox{$X \cap f(X \cup Y) \neq \emptyset$}.
By Contraction, \mbox{$f(X \cup Y) \subseteq$} \mbox{$X \cup Y$}
and, therefore, 
\[
f(X \cup Y) \subseteq X \cap f(X \cup Y) \cup Y \cap f(X \cup Y) =
X \cap f(X \cup Y), 
\]
by assumption.

Before we prove the last property needed, \ref{eq:union}, let us prove
a lemma.
\begin{lemma}
\label{le:funion}
\mbox{$f(\bigcup_{i \in I} X_{i}) \subseteq \bigcup_{i \in I}f(X_{i})$}.
\end{lemma}
\proof
By Coherence, we have
\mbox{$X_{i} \cap f(\bigcup_{i \in I} X_{i}) \subseteq$}
\mbox{$f(X_{i})$}.
By Contraction, 
\mbox{$f(\bigcup_{i \in I} X_{i}) \subseteq$}
\mbox{$\bigcup_{i \in I} X_{i}$}.
\QED
For property~\ref{eq:union}, assume \mbox{$f(X) \neq \emptyset$}
and \mbox{$Y_{i} \cap f(X \cup Y_{i}) = \emptyset$}
for any \mbox{$i \in I$}.
By Lemma~\ref{le:funion}, 
\mbox{$f(X \cup \bigcup_{i \in I} Y_{i}) \subseteq$}
\mbox{$\bigcup_{i \in I} f(X \cup Y_{i})$}.
But, by assumption, \mbox{$f(X \cup Y_{i}) \subseteq$} \mbox{$X$}, 
for any \mbox{$i \in I$}.
Therefore, for any \mbox{$j \in I$},
\mbox{$f(X \cup \bigcup_{i \in I} Y_{i}) \subseteq$}
\mbox{$X \cup Y_{j}$}.
By Coherence, then,
\mbox{$(X \cup Y_{j}) \cap f(X \cup \bigcup_{i \in I} Y_{i}) \subseteq$}
\mbox{$f(X \cup Y_{j}) \subseteq X - Y_{j}$}.
We have shown that, for any \mbox{$j \in I$},
\mbox{$f(X \cup \bigcup_{i \in I} Y_{i}) \subseteq$} \mbox{$X - Y_{j}$}.
We conclude that
\[
\bigcup_{j \in I} Y_{j} \cap f(X \cup \bigcup_{i \in I} Y_{i}) 
= \emptyset.
\]
\QED
Before we get to the second leg of our equivalence trip,
let us notice an additional property.
\begin{definition}
\label{def:heavy}
If \mbox{$x \in X$} we shall say that $x$ is {\em heavy} in $X$
iff \mbox{$X \not > \{ x \} $}.
\end{definition}
\begin{lemma}
\label{le:heavy}
If the relation $>$ is defined as in Theorem~\ref{the:chpl},
then, if \mbox{$f(X) = \emptyset$} all the members of $X$ are heavy,
and if \mbox{$f(X) \neq \emptyset$}, the heavy elements of $X$ are 
precisely the members of $f(X)$.
\end{lemma}
The proof is obvious.

The second half of the equivalence between choice functions and
qualitative measures will be described now.
\begin{theorem}
\label{the:plch}
Suppose $>$ is a qualitative measure that satisfies 
properties~\ref{eq:porder}, \ref{eq:inc}, \ref{eq:small}, \ref{eq:in}
and~\ref{eq:union}.
Then, the choice function $f$ defined by taking for
$f(X)$ the set of heavy elements (see Definition~\ref{def:heavy}) of $X$
satisfies Contraction, Coherence and Local Monotonicity.
If \cC\ is defined by Equation~\ref{eq:secdef}, it satisfies
Equation~\ref{eq:fdef}.
\end{theorem}
\proof
By definition, $f$ satisfies Contraction.
For Coherence, assume \mbox{$X \subseteq Y$} and \mbox{$x \in X$}
is heavy in $Y$, i.e., \mbox{$Y \not > \{ x \}$}.
By property~\ref{eq:inc}, \mbox{$X \not > \{ x \}$}
and $x$ is heavy in $X$.
We have proved that $f$ satisfies Coherence.

For Local Monotonicity, assume \mbox{$Y \subseteq X$}
and all heavy elements of $X$ are in $Y$.
Let $y$ be a heavy element of $Y$.
We know that \mbox{$y \in X$}.
We must show that $y$ is heavy in $X$.
Since all heavy elements in $X$ are in $Y$, 
any member $z$ of $X - Y$ is not heavy in $X$ 
and therefore \mbox{$X > \{ z \}$}. 
By property~\ref{eq:union}, we have
\mbox{$X > X - Y$}.
If $y$ was not heavy in $X$, again by~\ref{eq:union},
we would have \mbox{$X > (X - Y) \cup \{ y \}$}.
By property~\ref{eq:in}, then, 
\mbox{$Y - \{ y \} >$}
\mbox{$X - Y \cup \{ y \} $}.
By property~\ref{eq:inc}, then, we would have
\mbox{$Y > \{ y \}$}, contrary to the assumption
that $y$ is heavy in $Y$.
We have shown that $f$ satisfies Local Monotonicity.

We must now show that \cC\  satisfies Equation~\ref{eq:fdef}.
Our proof will be exactly the same as the corresponding part
of the proof of Theorem~\ref{the:chpl}, once we have shown 
that the property used as a definition
in Theorem~\ref{the:chpl} holds true.
\begin{equation}
\label{eq:central}
X > Y \ {\rm iff} \ f(X) \neq \emptyset \ 
{\rm and} \ Y \cap f(X \cup Y) = \emptyset 
\end{equation}
Assume \mbox{$X > Y$}.
By property~\ref{eq:inc}, \mbox{$X > \emptyset$}.
By property~\ref{eq:small}, there is some element of \mbox{$x \in X$}
such that \mbox{$\{ x \} > \emptyset$}.
The element $x$ is heavy in $\{x\}$ since $>$ is irreflexive, and therefore
\mbox{$x \in f(\{x\})$}.
By Coherence, we conclude that \mbox{$f(X) \neq \emptyset$}.
By property~\ref{eq:inc}, if \mbox{$y \in Y$}, then 
\mbox{$X > \{ y \}$} and $y$ is not heavy in $X$.
We have shown that \mbox{$Y \cap f(X) = \emptyset$}.

Suppose now that \mbox{$f(X) \neq \emptyset$} and
\mbox{$Y \cap f(X \cup Y) = \emptyset$}.
For any element \mbox{$y \in Y$},
\mbox{$X \cup Y > \{ y \}$}.
By property~\ref{eq:union},
\mbox{$ X \cup Y > Y$}.
By property~\ref{eq:in}, \mbox{$X > Y$}.
\QED
We have shown that the two very different semantic frameworks
proposed by choice functions and qualitative measures are
equivalent. This is a clear indication that the notion
captured is important and that the
operations \cC\ that may be defined in those frameworks
form a class of great interest.
Section~\ref{sec:nonmonded} proposes a characterization of those operations.
The representation result is proved in Section~\ref{sec:mainchar}.
\section{Choice functions: a general semantics for nonmonotonic
operations}
\label{sec:choice}
It is now time to get rid of the simplifying assumption
that all sets of models are definable (see Definition~\ref{def:definable}),
made in Section~\ref{subsec:simplechoice}.
In the general case, when certain sets of models are not definable,
we must take a second look at the properties of choice functions
considered in Section~\ref{subsec:simplechoice}.
Notice, first, that in the definition of $\cC(A)$ as
\mbox{$\overline{f(\widehat{A})}$}, the argument of $f$ is a definable
set ($\widehat{A}$). We have no need for applying $f$ to a set that
is not definable and,
therefore, we shall assume that the domain of $f$ is the set of
definable sets of models.
First, we shall require that the image under $f$ of a definable set
be definable.
This requirement has been introduced by Schlechta
in~\cite{Schle:91}.
Then, we shall understand the variables $X$ and $Y$ appearing in the
definition of the properties of $f$ that we considered
(Contraction, Coherence and Local Monotonicity), {\em not} as ranging
over all sets of models $X$ and $Y$, but only over {\em definable}
such sets.

On the first point,
we expect to be able to describe (by a set of formulas)
the sets in which we are interested, and on which we want to apply
the choice function $f$. But, similarly, we expect the result of the
application of $f$, the set of {\em preferred} elements of a set $X$
to be definable by a set of formulas (otherwise, how could we describe
it in the language at our disposal?).
We shall consider choice functions that are defined only on definable sets 
of models and that send definable sets to definable sets, i.e., we assume:
\[
{\bf Definability \ Preservation} \ \ \forall A \subseteq \cL , \ 
f(\widehat{A}) = \widehat{\overline{f(\widehat{A})}}
\]
The identity function obviously preserves definability.
This property has never been considered by the Social Choice community,
which seems to have been interested so far only in the case that 
\cM\ is finite.

On the second point, for example, Local Monotonicity is now understood as:
for any {\em definable} sets $X$, $Y$, such that 
\mbox{$f(X) \subseteq$} \mbox{$Y \subseteq$} $X$, one has
\mbox{$f(Y) \subseteq f(X)$}.
It turns out (the proof is left to the reader)
that one may extend any $f$ defined on definable sets
(and satisfying Contraction, Coherence and Local Monotonicity)
to arbitrary subsets by:
\mbox{$f(X) = X \cap f(\widehat{\overline{X}})$}.
This extension satisfies Contraction and Coherence for any subsets
$X$, $Y$, but it does not satisfy
Local Monotonicity for arbitrary such sets.
It satisfies the following: for any sets $X$, $Y$ 
such that \mbox{$f(\widehat{\overline{X}}) \subseteq Y \subseteq X$},
one has \mbox{$f(Y) \subseteq f(X)$}.

In the next section, the family of nonmonotonic operations defined
by Equation~\ref{eq:fdef} from definability-preserving 
choice functions that satisfy
Contraction, Coherence and Local Monotonicity will
be described.
\section{Representation result}
\label{sec:mainchar}
We shall now state and then prove our main characterization
result. It proves an exact correspondence between the choice function
semantics of Section~\ref{sec:choice} 
and the properties of \cC\  described in Section~\ref{sec:nonmonded}.
Theorems~\ref{the:chpl} and~\ref{the:plch} show that those properties
also correspond exactly with
the qualitative
measures of Section~\ref{subsec:qualmeas}, at least under the simplifying
assumption of Section~\ref{subsec:simple}.
Lifting this simplifying assumption there would involve a careful
study of which sets of models are {\em measurable} in the qualitative 
sense. This can surely be done.
\begin{theorem}
\label{the:main}
Suppose we are given a language \cL \ and a function 
\mbox{$\cC : 2^{\cL} \longrightarrow 2^{\cL}$}.
Then, the following two conditions are equivalent:
\begin{enumerate}
\item \label{C}
\cC\  satisfies Inclusion, Idempotence, Cautious Monotonicity, Conditional Monotonicity
and Threshold Monotonicity,
\item \label{f}
there exists a set \cM \ (of models), a satisfaction relation
\mbox{$\models \: \subseteq \cM \times \cL$} and a definability-preserving
choice function
\mbox{$f : D_{\cM} \longrightarrow D_{\cM}$} 
satisfying Contraction, Coherence and Local Monotonicity such that
\mbox{$\cC(A) = \overline{f(\widehat{A}})$}. 
\end{enumerate}
\end{theorem}
The notation $D_{\cM}$ is explained in Definition~\ref{def:definable}.
\proof
First, let us show soundness: property~\ref{f} implies property~\ref{C}.
Assume \cM\ is a set and $\models$ a binary relation on 
\mbox{$\cM \times \cL$}.
Assume $f$ satisfies Definability Preservation, 
Contraction, Coherence and Local Monotonicity.
Define \mbox{$\cC(A) = \overline{f(\widehat{A})}$}.
We shall show that \cC \  satisfies Inclusion, Idempotence, 
Cautious Monotonicity, Conditional Monotonicity and Threshold Monotonicity.

For Inclusion, notice that, by Contraction, 
\mbox{$f(\widehat{A}) \subseteq$}
\mbox{$\widehat{A}$}, and therefore
\mbox{$\overline{\widehat{A}} \subseteq$}
\mbox{$\overline{f(\widehat{A})} =$}
\mbox{$\cC(A)$}.
But \mbox{$A \subseteq \overline{\widehat{A}}$} and we conclude
that \mbox{$A \subseteq \cC(A)$}.
We shall now prove a lemma that makes use of the definability
preservation property.
\begin{lemma}
\label{le:one}
\mbox{$f(\widehat{A}) = \widehat{\cC(A)}$}.
\end{lemma}
\proof
By Definability Preservation,
\mbox{$f(\widehat{A}) =$}
\mbox{$\widehat{\overline{f(\widehat{A})}}$}.
But \mbox{$\widehat{\overline{f(\widehat{A})}} =$}
\mbox{$\widehat{\cC(A)}$}.
\QED
For Idempotence, notice that, by Inclusion (already proved), we have
\mbox{$A \subseteq \cC(A)$} and therefore
\mbox{$\widehat{\cC(A)} \subseteq \widehat{A}$}.
By Coherence, then, we have
\mbox{$\widehat{\cC(A)} \cap f(\widehat{A}) \subseteq$}
\mbox{$f(\widehat{\cC(A)})$}.
By Lemma~\ref{le:one}, 
\mbox{$\widehat{\cC(A)} \cap f(\widehat{A}) =$}
\mbox{$f(\widehat{A})$}
and \mbox{$f(\widehat{A}) \subseteq$}
\mbox{$f(\widehat{\cC(A)})$}.
We conclude that we have \mbox{$\cC(\cC(A)) \subseteq \cC(A)$}.
The opposite inclusion follows from Inclusion.

For Cautious Monotonicity, we use Local Monotonicity and Lemma~\ref{le:one}.
Assume \mbox{$A \subseteq$}
\mbox{$B \subseteq \cC(A)$}.
We have \mbox{$\widehat{\cC(A)} \subseteq$}
\mbox{$\widehat{B} \subseteq \widehat{A}$}.
By Lemma~\ref{le:one},
\mbox{$f(\widehat{\overline{\widehat{A}}}) =$}
\mbox{$f(\widehat{A}) \subseteq$}
\mbox{$\widehat{B} \subseteq \widehat{A}$}.
Local Monotonicity, then, implies that
\mbox{$f(\widehat{B}) \subseteq$}
\mbox{$f(\widehat{A})$}.
We conclude that \mbox{$\cC(A) \subseteq \cC(B)$}.

The next remark will be useful.
\begin{lemma}
\label{le:two}
\mbox{$\overline{\widehat{A}} \subseteq \cC(A)$}.
\end{lemma}
\proof
Since, by Contraction, 
\mbox{$f(\widehat{A}) \subseteq \widehat{A}$}.
\QED
We shall prove Conditional Monotonicity and Threshold Monotonicity. 
Since \mbox{$\widehat{A \cup B} \subseteq$}
\mbox{$\widehat{A}$},
Coherence implies that we have
\mbox{$\widehat{A \cup B} \cap f(\widehat{A}) \subseteq$}
\mbox{$f(\widehat{A \cup B})$}.
By Lemma~\ref{le:one},
\mbox{$\widehat{A \cup B} \cap \widehat{\cC(A)} \subseteq$}
\mbox{$f(\widehat{A \cup B})$}.
Therefore \mbox{${\rm Mod}(A \cup B \cup \cC(A)) \subseteq$}
\mbox{$f(\widehat{A \cup B})$} and
\[
\cC(A , B) \subseteq \overline{{\rm Mod}(A \cup B \cup \cC(A))} =
\overline{{\rm Mod}(\cC(A) \cup B)} \subseteq 
\overline{{\rm Mod}(\cC(A) \cup B \cup C)}, \forall C.
\]
By Lemma~\ref{le:two}, then,
\mbox{$\cC(A , B) \subseteq$}
\mbox{$\cC(\cC(A) , B , C)$}.
Conditional Monotonicity is obtained by taking 
\mbox{$C = \emptyset$}.
For Threshold Monotonicity, take \mbox{$A = \cC(D)$} and
use Idempotence.
 
We have proved the soundness part.
Let us proceed to the proof of completeness: property~\ref{C} 
implies property~\ref{f}.
Assume that \cC\  satisfies Inclusion, Idempotence, Cautious Monotonicity,
Conditional Monotonicity and Threshold Monotonicity.
We shall now describe \cM\ and the satisfaction relation $\models$
in a unsurprising way.
We shall take \cM\ to be set of all sets $T$ of formulas
such that \mbox{$T = \cC(T)$}.
Such a set $T$ will be called a theory.
We shall define $\models$ by:
\mbox{$T \models a$} iff \mbox{$a \in T$}.

Let us draw some consequences from the definition of $\models$.
\begin{lemma}
\label{le:hat}
For any \mbox{$A \subseteq \cL$}, 
$\widehat{A}$ is the set of all theories that include $A$ and
\mbox{$\overline{\widehat{A}} =$}
\mbox{$\bigcap_{B \subseteq \cL} \cC(A , B)$}.
\end{lemma}
\proof
By definition, $\widehat{A}$ is the set of all theories that include $A$
and $\overline{\widehat{A}}$ is the intersection of all theories
that include $A$.
Let $T$ be any theory that includes $A$: 
\mbox{$A \subseteq$}
\mbox{$T =$}
\mbox{$\cC(T) =$}
\mbox{$\cC(A , T)$}
and therefore \mbox{$T = \cC(A , B)$} for some $B$.
We have shown that 
\mbox{$\bigcap_{B \subseteq \cL} \cC(A , B) \subseteq$}
\mbox{$\overline{\widehat{A}}$}.

But, \mbox{$\cC(A , B)$} is a theory, by Idempotence, and it includes
$A$ by Inclusion. Therefore
\mbox{$\overline{\widehat{A}} \subseteq$}
\mbox{$\bigcap_{B \subseteq \cL} \cC(A , B)$} and our claim is proved. 
\QED
To simplify notations we shall write:
\begin{equation}
\label{eq:Cn}
\Cn(A) = \overline{\widehat{A}} = \bigcap_{B \subseteq \cL} \cC(A , B).
\end{equation}
Since $\Cn(A)$ is the intersection of the
sets $\cC(B)$ for all sets $B$ including $A$,
it is the largest monotonic sub-mapping of \cC.
Since \mbox{$\Cn(A) = \overline{\widehat{A}}$}, it is
a consequence operation, i.e., satisfies Inclusion, Idempotence
and Monotonicity.
This remark is very close to a result of J. Dietrich in~\cite{Dietrich:94}.
Notice that, by Conditional Monotonicity and Threshold Monotonicity
we have:
\begin{equation}
\label{eq:cCCncC}
\cC(A , B) \subseteq \Cn(\cC(A) , B).
\end{equation}
\begin{lemma}
\label{le:ACNC}
\mbox{$A \subseteq \Cn(A) \subseteq \cC(A)$}.
\end{lemma}
\proof
We have seen that \Cn\  satisfies Inclusion.
By taking, in the definition of \Cn, \mbox{$B = \emptyset$},
one sees that \mbox{$\Cn(A) \subseteq \cC(A)$}.
\QED
\begin{lemma}[Right Absorption]
\label{le:RA}
\cC(\Cn(A)) = \cC(A).
\end{lemma}
\proof
By Lemmas~\ref{le:ACNC} and~\ref{le:cum}.
\QED
\begin{lemma}[Left Absorption]
\label{le:LA}
\Cn(\cC(A)) = \cC(A).
\end{lemma}
\proof
By Lemma~\ref{le:ACNC}, we have 
\mbox{$\cC(A) \subseteq \Cn(\cC(A)) \subseteq \cC(\cC(A))$}.
By Idempotence, we have
\mbox{$\cC(A) = \Cn(\cC(A)) = \cC(\cC(A))$}.
\QED
We may now define the choice function $f$.
Consider an arbitrary definable set \mbox{$X = \widehat{A} \subseteq \cM$}.
By Lemma~\ref{le:RA}, \mbox{$\widehat{A} = \widehat{B}$} implies
\mbox{$\cC(A) = \cC(B)$} and therefore we may 
define \mbox{$f(X)$} by:
\begin{equation}
\label{eq:efdef}
f(\widehat{A}) = \widehat{\cC(A)}.
\end{equation}
One immediately sees that the choice function $f$ preserves definability,
since \mbox{$\widehat{\cC(A)}$} is definable.
We must now show that $f$ satisfies Contraction, Coherence and Local
Monotonicity, and that 
\begin{equation}
\label{eq:m}
\cC(A) = \overline{f(\widehat{A})}.
\end{equation}
Let us deal with this last question first.
By~(\ref{eq:efdef}),
\mbox{$f(\widehat{A}) = \widehat{\cC(A)}$}.
Therefore 
\mbox{$\overline{f(\widehat{A})} =$}
\mbox{$\overline{\widehat{\cC(A)}} =$}
\mbox{$\Cn(\cC(A))$}.
By Lemma~\ref{le:LA},
\mbox{$\overline{f(\widehat{A})} = \cC(A)$}.
We have shown that Equation~\ref{eq:m} holds.

It is clear from Equation~\ref{eq:efdef} that $f$ satisfies Contraction.
Let us prove now that $f$ satisfies Coherence.
Assume \mbox{$X \subseteq Y$}.
We have \mbox{$\overline{Y} \subseteq \overline{X}$} and therefore
\mbox{$\cC(\overline{X}) =$} \mbox{$\cC(\overline{Y} , \overline{X})$}.
By Equation~\ref{eq:cCCncC}, we have
\mbox{$\cC(\overline{X}) \subseteq$}
\mbox{$\Cn(\cC(\overline{Y}) , \overline{X})$}.
Therefore, \mbox{${\rm Mod}(\cC(\overline{Y}) \cup \overline{X}) \subseteq$}
\mbox{${\rm Mod}(\cC(\overline{X}))$} and
\mbox{$\widehat{\cC(\overline{Y})} \cap \widehat{\overline{X}} \subseteq$}
\mbox{$\widehat{\cC(\overline{X})}$}.
But, \mbox{$X \subseteq \widehat{\overline{X}}$} and
therefore
\[
X \cap f(Y) = X \cap Y \cap \widehat{\cC(\overline{Y})} \subseteq
\widehat{\overline{X}} \cap \widehat{\cC(\overline{Y})} \subseteq
\widehat{\cC(\overline{X})}.
\]
We conclude that 
\mbox{$ X \cap f(Y) \subseteq$}
\mbox{$X \cap \widehat{\cC(\overline{X})} =$}
\mbox{$f(X)$}.

Finally, let us show that $f$ satisfies Local Monotonicity.
Assume 
\mbox{$f(\widehat{\overline{X}}) \subseteq$}
\mbox{$Y \subseteq X$}.
We have, by Equation~\ref{eq:m}, 
\mbox{$\overline{X} \subseteq$}
\mbox{$\overline{Y} \subseteq$}
\mbox{$\overline{f(\widehat{\overline{X}})} =$}
\mbox{$\cC(\overline{X})$}.
By Cautious Monotonicity, then, we conclude that
\mbox{$\cC(\overline{X}) \subseteq$}
\mbox{$\cC(\overline{Y})$} and
\mbox{$\widehat{\cC(\overline{Y})} \subseteq$}
\mbox{$\widehat{\cC(\overline{X})}$}.
Since \mbox{$Y \subseteq X$},
\mbox{$Y \cap \widehat{\cC(\overline{Y})} \subseteq$}
\mbox{$X \cap \widehat{\cC(\overline{X})}$}.
The proof of Theorem~\ref{the:main} is now complete.
\QED
One may check that, if the operation \cC\ is monotonic, then,
the operation $f$ defined in the construction above is the identity
on definable sets: \mbox{$f(\widehat{A})$} is \mbox{$\widehat{\cC(A)}$},
the set of all theories that include \mbox{$\cC(A)$}, which is equal to
$\widehat{A}$, the set of all theories that include $A$.
\section{Properties of nonmonotonic operations}
\label{sec:Properties}
\subsection{First Properties}
\label{subsec:firstp}
We shall consider some properties of the operations that
satisfy Inclusion, Idempotence, Cautious Monotonicity,
Conditional Monotonicity and Threshold Monotonicity, {\em nonmonotonic}
operations in the sequel.

For such operations, the intersection of two theories is not always
a theory. 
Consider, for example, the language that contains two
elements $a$ and $b$ and the operation defined by:
\mbox{$\cC(A) = A$} for any \mbox{$A \neq \emptyset$}
and \mbox{$\cC(\emptyset) = \{ a \}$}.
This is not a monotonic operation: the only breach of monotonicity
is \mbox{$\cC(\emptyset) \not \subseteq \cC(b)$}.
It is easy to check that it satisfies our conditions.
But \mbox{$\cC(a) \cap \cC(b) = \emptyset$}
is not a theory.

In the monotonic framework, the operation \cC\ is defined
by the set of its theories, \mbox{$\cC(A)$} being the intersection
of all theories including $A$.
In the nonmonotonic framework, this is not the case.
Two different operations may define the same set of theories.
The example above will prove our case.
The operation $\cC'$ defined by \mbox{$\cC'(\emptyset) = \{ b \}$}
and otherwise \mbox{$\cC'(X) = \cC(X)$} is different from \cC\
but it has exactly
the same theories as \cC.

The property concerning the intersection of a family of operations
holds, but its proof is more intricate that in the monotonic case.
If $\cC_{i}$ is a family of nonmonotonic operations (i.e., satisfying
Inclusion, Idempotence, Cautious Monotonicity, Conditional Monotonicity
and Threshold Monotonicity),
its intersection is a nonmonotonic operation.
The easiest proof may be semantic: each $\cC_{i}$
is defined by a set $\cM_{i}$ and a choice function
$f_{i}$.
Consider the set \mbox{$\bigcup_{i} \cM_{i}$} (assume the $\cM_{i}$
have pairwise empty intersections) and
the function $f$ that operates as $f_{i}$ on $\cM_{i}$ and
takes the union of the sets obtained this way.
It is easy to see it satisfies Inclusion, Coherence and
Local Monotonicity. The operation defined this way is
the intersection of the $\cC_{i}$'s.

It is easy to see that, if \cC\ is nonmonotonic, i.e., satisfies
Inclusion, Idempotence, Cautious Monotonicity, Conditional Monotonicity and
Threshold Monotonicity, then,
so is $\cC'$ defined by: \mbox{$\cC'(A) = \cC(A , B)$}.
\subsection{Connectives}
\label{subsec:conn}
We shall now find elegant properties of (nonmonotonic) \cC\ 
that characterize the (semantically)
classical propositional connectives.
We shall see that the nonmonotonic logic of the (semantically)
classical connectives 
is {\em classical} propositional logic.
We do not consider, in this work, the many possibilities offered
by nonmonotonic logics for the definition of non-classical connectives,
i.e., connectives that are not defined by truth tables.
As in the case of monotonic logic, some additional 
compactness property is needed for the completeness result.
The property required is weak and does not amount at all to requiring
the operation \cC\ to be compact. 

A word of caution about terminology is needed here.
In sequent calculus presentations, connectives are characterized by
left and right rules.
In natural deduction presentations, they are characterized by introduction
and elimination rules.
We mix those two terminologies freely to name the rules we are interested
in.

Let us consider, first, the case of conjunction.
Assume that \cL\ is closed under a binary $\wedge$ and that 
we consider only satisfaction relations that satisfy:
\begin{equation}
\label{eq:and}
x \models a \wedge b \ {\rm iff} \ x \models a \ {\rm and} \  x \models b.
\end{equation}
Then \cC, defined by \mbox{$\cC(A) = \overline{f(\widehat{A})}$}
satisfies:
\begin{equation}
\label{eq:andrule}
\cC(A , a \wedge b) = \cC(A , a , b).
\end{equation}
The proof is very easy: 
the models that satisfy $A$ and $a \wedge b$ are exactly those that 
satisfy $A$, $a$ and $b$.
Notice that the treatment of conjunction is exactly the same
as in the monotonic case.

Let us, now, consider the case of negation.
Assume \cL\ is closed under a unary $\neg$ and that 
we consider only satisfaction relations that satisfy:
\begin{equation}
\label{eq:neg}
x \models \neg a \ {\rm iff} \ x \not \models a .
\end{equation}
The left introduction rule of the monotonic case:
\mbox{$a \in \cC(A)$} implies \mbox{$\cC(A , \neg a) =$} \mbox{$\cL$}
is not valid in our nonmonotonic framework.
The rules we propose to characterize negation are the following.
\begin{equation}
\label{eq:negleftintro}
\cC(A , a , \neg a) = \cL
\end{equation}
\begin{equation}
\label{eq:negleftelim}
\cC(A , \neg a) = \cL \Rightarrow a \in \cC(A)
\end{equation}
The validity of those rules is easy to prove.
For the first one, there are no models that satisfy 
\mbox{$A \cup \{a\} \cup \{\neg a\}$},
therefore, by Contraction we have
\mbox{$f({\rm Mod}(A \cup \{a\} \cup \{\neg a\})) =$} \mbox{$\emptyset$}
and \mbox{$\cC(A , a , \neg a) =$} \mbox{$\cL$}.
For the second one, since no model satisfies all formulas,
if \mbox{$\cC(A , \neg a) =$} \mbox{$\cL$}, it must be the case that
\mbox{$f({\rm Mod}(A \cup \{ \neg a \})) =$} \mbox{$\emptyset$}.
By Coherence, \mbox{${\rm Mod}(\{ \neg a\} ) \cap f(\widehat{A}) = \emptyset$}.
Therefore \mbox{$f(\widehat{A}) \subseteq$} \mbox{${\rm Mod}(\{ a \})$} and
\mbox{$a \in \cC(A)$}.
One may notice that a similar result fails in Relevance Logic~\cite{Restall:NegRel},
where the {\em Boolean negation} of~\cite{Meyer:74}, defined by (\ref{eq:neg}),
is not reasonable.

Consider, now, the case of disjunction.
Assume \cL\ is closed under a binary $\vee$ and that 
we consider only satisfaction relations that satisfy:
\begin{equation}
\label{eq:or}
x \models a \vee b \ {\rm iff \ either} \ x \models a \ {\rm or} \ x \models b.
\end{equation}
The Or introduction rule of monotonic sequent calculus, 
a left introduction rule, is valid:
\begin{equation}
\label{eq:orleftintro}
\cC(A , a) \cap \cC(A , b) \subseteq \cC(A , a \vee b).
\end{equation}
This follows easily from the fact that a model that satisfies
$a \vee b$ satisfies at least one of $a$ or $b$ and from Coherence:
\[
f({\rm Mod}(A \cup \{a \vee b\})) \subseteq
f(\widehat{A} \cap \widehat{\{ a \}}) 
\cup f(\widehat{A} \cap \widehat{\{ b \}}).
\]
The right elimination rule of the monotonic case:
\mbox{$\cC(A , a \vee b) \subseteq \cC(A , a) \cap \cC(A , b)$}
is not valid. We replace it by a right introduction rule.
\begin{equation}
\label{eq:orrightintro}
a \in \cC(A) \: \Rightarrow \: a \vee b \in \cC(A) , \:
b \in \cC(A) \: \Rightarrow \: a \vee b \in \cC(A) . 
\end{equation} 
Its validity is obvious.

Lastly, consider the case of material implication.
Assume \cL\ is closed under a binary $\rightarrow$ and that 
we consider only satisfaction relations that satisfy:
\begin{equation}
\label{eq:imp}
x \not \models a \rightarrow b \ {\rm iff} \ x \models a \ {\rm and} \ 
 x \not \models b.
\end{equation}
The right introduction rule of the monotonic case is valid:
\begin{equation}
\label{eq:imprightintro}
b \in \cC(A , a) \: \Rightarrow \: a \rightarrow b \in \cC(A).
\end{equation}
Assume \mbox{$b \in \cC(A , a)$}.
Let $B$ be set of all models in $f(\widehat{A})$ that satisfy $a$.
By Coherence, \mbox{$B \subseteq f({\rm Mod}(A \cup \{ a \}))$} 
and therefore all models in $B$ satisfy $b$ and also
\mbox{$a \rightarrow b$}.
All models in \mbox{$f(\widehat{A}) - B$} satisfy 
\mbox{$a \rightarrow b$}.
We conclude that all models in $f(\widehat{A})$ satisfy
\mbox{$a \rightarrow b$} and
\mbox{$a \rightarrow b \in \cC(A)$}.
The right elimination rule of the monotonic case:
\mbox{$a \rightarrow b \in \cC(A)$} implies 
\mbox{$b \in \cC(A , a)$} is not valid.
We shall use the following left introduction rule:
\begin{equation}
\label{eq:impleftintro}
b \in \cC(A , a , a \rightarrow b)
\end{equation}
which is easily seen to be valid.

Suppose \cL\ is closed under some subset of the connectives
$\wedge$, $\neg$, $\vee$ and $\rightarrow$.
Is it the case that any operation \cC\ that satisfies
Inclusion, Idempotence, Cautious Monotonicity, Conditional Monotonicity,
Threshold Monotonicity and the two properties corresponding to
each one of the connectives concerned may be defined by a 
set of models, a satisfaction relation that satisfies the requirement 
concerning each of the connectives considered and a
choice function that satisfies Inclusion, Coherence and Local
Monotonicity?

The answer cannot be positive in general. 
Although the answer is positive for a language \cL\ that contains
conjunction as its sole connective, it is negative
as soon as \cL\  contains a negation.
A compactness assumption will ensure a positive answer.
Without any such assumption, already in the monotonic framework, 
the result does not hold.
Consider the following example. The language \cL\ has
an infinite set of atomic propositions and is closed under
negation.
The monotonic operation \cC\  satisfies 
\mbox{$a \in \cC(A)$} iff \mbox{$\cC(A , \neg a) =$} \mbox{$\cL$},
and therefore we may always remove double negations and is defined
(up to removal of double negations) by:
\mbox{$\cC(A) = A$}
if $A$ is finite and does not contain an atomic proposition and
its negation, and \mbox{$\cC(A) = \cL$} otherwise (i.e., if $A$ is infinite
or contains an atomic proposition and its negation).
Notice that this \cC\  fails the Lindenbaum lemma:
there are consistent sets but no maximal consistent set.
Assume \cC\ is representable by a suitable $f$.
Let \mbox{$Y = f(\widehat{\emptyset})$} and \mbox{$y \in Y$}.
We have \mbox{$\{y\} = \widehat{\overline{y}}$}, and,
by Coherence, \mbox{$y \in f(\{y\})$}.
Also,
\mbox{$f(\{y\}) = \widehat{\cC(\overline{y})}$}.
But $\overline{y}$ includes a infinite number of atomic propositions
or an infinite number of negations of atomic propositions,
therefore \mbox{$\cC(\overline{y}) = \cL$} and \mbox{$f(\{y\}) = \emptyset$}.
A contradiction to \mbox{$y \in f(\{y\})$}. 
We conclude that \mbox{$Y = \emptyset$}.
But then \mbox{$\cC(\emptyset) = \cL$}, a contradiction.

We shall, then, assume that \cC\  satisfies the following:
\[
{\bf Weak \ Compactness} \ \ \ \ \cC(A) = \cL \: \Rightarrow \:
\exists {\rm \ a \ finite \ } B \subseteqf A \ {\rm such \ that} \ 
\cC(B) = \cL.
\]
Notice that the Weak Compactness assumed does not imply, in the nonmonotonic
framework, that, if \mbox{$a \in \cC(A)$}, then there is a finite
subset $B$ of $A$ such that \mbox{$a \in \cC(B)$}, even when proper
connectives are available.
In the monotonic case, the monotonic left introduction rule for negation
makes Compactness follow from Weak Compactness, but we rejected this rule. 

Before we state and prove the main theorem of this section,
let us build the tools for the proof.
We shall assume that \cC\  satisfies Inclusion, Idempotence,
Cautious Monotonicity, Conditional Monotonicity, Threshold Monotonicity
and Weak Compactness.
We assume that \cL\ is closed under some set of
connectives that includes negation and that \cC\  satisfies
the two properties described above for
each of the connectives assumed to be in the language
(in particular it satisfies~\ref{eq:negleftintro} and~\ref{eq:negleftelim}).
Recall that $A$ is a theory iff \mbox{$A = \cC(A)$}.
\begin{definition}
\label{def:incon}
A set \mbox{$A \subseteq \cL$} is said to be {\em inconsistent}
iff \mbox{$\cC(A) = \cL$}.
A set that is not inconsistent is said to be {\em consistent}.
\end{definition}
Notice that, by Idempotence, $A$ is consistent iff $\cC(A)$ is
consistent, and that there is only one inconsistent theory, namely \cL.
\begin{lemma}
\label{le:incon}
If \mbox{$A \subseteq B$} and $A$ is inconsistent,
then $B$ is inconsistent.
\end{lemma}
\proof
We have: \mbox{$A \subseteq B \subseteq \cL = \cC(A)$}.
By Cautious Monotonicity, then, \mbox{$\cL =$}
\mbox{$\cC(A) \subseteq$}
\mbox{$\cC(B)$}.
\QED
The following notion is fundamental.
\begin{definition}
\label{def:maximal}
A set $A$ is said to be {\em maximal consistent} iff it is consistent
and any strict superset \mbox{$B \supset A$} is inconsistent.
\end{definition}
The next two lemmas are central.
\begin{lemma}
\label{le:Zorn}
If $A$ is consistent, there is some maximal consistent set $B$
such that \mbox{$A \subseteq B$}.
\end{lemma}
\proof
The proof is as in the classical case. It is included only for
completeness sake.
Consider any ascending chain of consistent sets, $A_{i}$,
\mbox{$i \in I$},
where \mbox{$i < j$} implies \mbox{$A_{i} \subseteq A_{j}$}.
We claim that the union \mbox{$B = \bigcup_{i \in I} A_{i}$}
is consistent.
If it were inconsistent, 
by Weak Compactness, there would be some finite inconsistent subset 
of $B$. This subset would be a subset of $A_{i}$ for some $i$,
and by Lemma~\ref{le:incon}, $A_{i}$ would be inconsistent, contrary
to assumption.
We have shown that the union of any ascending chain of consistent sets
is consistent.
Zorn's lemma, then, implies that any consistent $A$ may be embedded
in a maximal consistent set.
\QED
\begin{lemma}
\label{le:maxcons}
If $A$ is maximal consistent, then 
\begin{enumerate}
\item $A$ is a theory,
\item (if $\wedge$ is in the language) \mbox{$a \wedge b \in A$}
iff \mbox{$a \in A$} and \mbox{$b \in A$},
\item \mbox{$\neg a \in A$} iff \mbox{$a \not \in A$},
\item (if $\vee$ is in the language) \mbox{$a \vee b \in A$}
iff \mbox{$a \in A$} or \mbox{$b \in A$},
\item (if $\rightarrow$ is in the language) \mbox{$a \rightarrow b \not \in A$}
iff \mbox{$a \in A$} and \mbox{$b \not \in A$}.
\end{enumerate}
\end{lemma}
\proof
Assume $A$ is maximal consistent.
If $\cC(A)$ were a strict superset of $A$ it would be inconsistent,
but then $A$ would be inconsistent.
Therefore \mbox{$\cC(A) = A$}.

By~\ref{eq:andrule}, and since $A$ is a theory,
\mbox{$a \wedge b \in A$} iff \mbox{$a \in A$} and \mbox{$b \in A$}.
The maximality of $A$ is not used here.

Assume \mbox{$\neg a \in A$}. If we had \mbox{$a \in A$}, we would, 
by~\ref{eq:negleftintro} have \mbox{$\cC(A) = \cL$}, a contradiction
to the consistency of $A$.
Assume, now, \mbox{$\neg a \not \in A$}.
Since $A$ is maximal consistent, \mbox{$\cC(A , \neg a) = \cL$}.
By~\ref{eq:negleftelim}, we have \mbox{$a \in \cC(A) = A$}.

Assume \mbox{$a \in A$} or \mbox{$b \in A$}.
By the right introduction rule~\ref{eq:orrightintro},
\mbox{$a \vee b \in \cC(A) = A$}.
Assume, now that \mbox{$a \vee b \in A$}.
We must show that at least one of $a$ or $b$ is in $A$.
Suppose, a contrario, that neither one is in $A$.
Since $A$ is maximal consistent, both \mbox{$\cC(A , a)$} and
\mbox{$\cC(A , b)$} are equal to \cL.
But, then, the left introduction rule~\ref{eq:orleftintro}
implies \mbox{$\cC(A) =$}
\mbox{$\cC(A , a \vee b) = \cL$},
a contradiction.

Assume \mbox{$a \not \in A$}.
Since $A$ is maximal consistent, \mbox{$\cC(A , a) = \cL$},
and therefore \mbox{$b \in \cC(A , a)$} and, by the right introduction
rule~\ref{eq:imprightintro}, \mbox{$a \rightarrow b \in \cC(A) = A$}.
Suppose, now, that \mbox{$a \in A$}.
If \mbox{$a \rightarrow b \in A$}, then, by the left introduction
rule~\ref{eq:impleftintro}, \mbox{$b \in \cC(A) = A$}.
If \mbox{$a \rightarrow b \not \in A$}, then, by the right introduction
rule~\ref{eq:imprightintro}, \mbox{$b \not \in \cC(A , a) = \cC(A) = A$}.
\QED
The next theorem shows that the Introduction Elimination rules
above define exactly the (semantically) classical connectives.
\begin{theorem}
\label{the:propcalc}
Assume that \cL\ is closed under some of the propositional
connectives, including negation, and that \cC\  satisfies
the Introduction and Elimination properties described above for
each of the connectives of the language.
Let \cC\  satisfy Inclusion, Idempotence,
Cautious Monotonicity, Conditional Monotonicity, Threshold Monotonicity
and Weak Compactness.
Then, there is a set \cM, a satisfaction relation $\models$ {\em that
behaves classically} for each of the existing connectives and
a definability-preserving choice function $f$ that satisfies
Contraction, Coherence and Local Monotonicity that defines \cC, i.e., 
such that
\mbox{$\cC(A) = \overline{f(\widehat{A})}$}, for any \mbox{$A \subseteq \cL$}.
\end{theorem}
\proof
The proof proceeds exactly as the proof of the completeness part
of Theorem~\ref{the:main}, except that, for the set \cM\ we take, 
not all theories, but only the maximal consistent sets of formulas.
The proof proceeds exactly in the same way, as soon as we have proved
Lemma~\ref{le:repl} to replace Lemma~\ref{le:hat}.
The fact that the satisfaction relation behaves classically for the
connectives follows from Lemma~\ref{le:maxcons}.
\QED
\begin{lemma}
\label{le:repl}
For any \mbox{$A \subseteq \cL$}, 
$\widehat{A}$ is the set of all maximal consistent sets that include $A$ and
\mbox{$\overline{\widehat{A}} =$}
\mbox{$\bigcap_{B \subseteq \cL} \cC(A , B)$}.
\end{lemma}
\proof
By definition, $\widehat{A}$ is the set of all maximal consistent
sets that include $A$
and $\overline{\widehat{A}}$ is the intersection of all maximal consistent
sets that include $A$.
Let $T$ be such a set.
By Lemma~\ref{le:maxcons}, $T$ is a theory and: 
\mbox{$A \subseteq$}
\mbox{$T =$}
\mbox{$\cC(T) =$}
\mbox{$\cC(A , T)$}
and therefore \mbox{$T = \cC(A , B)$} for some $B$.
We have shown that 
\mbox{$\bigcap_{B \subseteq \cL} \cC(A , B) \subseteq$}
\mbox{$\overline{\widehat{A}}$}.

But, suppose, now, that 
\mbox{$a \not \in \cC(A , B)$}.
By~\ref{eq:negleftelim}, \mbox{$\cC(A , B , \neg a) \neq \cL$}
(we need here the assumption that negation is in the language).
By Lemma~\ref{le:Zorn},
\mbox{$\cC(A , B , \neg a)$} 
is a subset of some maximal consistent set.
This set, since it is consistent and contains $\neg a$, does not contain $a$
(see~\ref{eq:negleftintro}).
Therefore \mbox{$a \not \in \overline{\widehat{A}}$}.
We have shown that 
\mbox{$\overline{\widehat{A}} \subseteq$}
\mbox{$\bigcap_{B \subseteq \cL} \cC(A , B)$} and our claim is proved. 
\QED
One may ask whether the result holds even for languages that do not
include negation. The question is open.
But, notice that, in the proof, we make use of the fact that, if
$T$ is a theory and \mbox{$a \not \in T$}, then there is a maximal consistent
superset of $T$ that does not include $a$.
This does not hold in general: for example, if \cC\ is the operation
of deduction of intuitionistic logic and $p$ is an atomic proposition,
\mbox{$p \not \in \cC(\neg \neg p)$}, but any maximal consistent set
that includes $\neg \neg p$ must include $p$.

We may now show that propositional nonmonotonic
logic is not weaker than (and therefore exactly the same as) monotonic
logic.
\begin{theorem}
\label{the:prop}
Let \cL\  be a propositional calculus and \mbox{$a , b \in \cL$}.
The following propositions are equivalent.
\begin{enumerate}
\item \label{logimpl}
$a$ logically implies $b$, i.e., \mbox{$a \models b$},
\item \label{mon}
for any operation \cC\ that satisfies Inclusion, Idempotence, Monotonicity,
Weak Compactness and the 
Introduction-Elimination rules above:
\mbox{$b \in \cC(a)$},
\item \label{si}
for any operation \cC\ that satisfies Inclusion, Idempotence, Cautious
Monotonicity, Conditional Monotonicity and Threshold Monotonicity, Weak Compactness 
and the 
Introduction-Elimination rules above:
\mbox{$b \in \cC(a)$},
\item \label{forall}
for any such \cC\ and for any \mbox{$A \subseteq \cL$}:
\mbox{$b \in \cC(A , a)$},
\item \label{contra}
for any such \cC:
\mbox{$\cC(a , \neg b) = \cL$}.
\end{enumerate}
\end{theorem}
\proof
Property~\ref{contra} implies~\ref{forall},
since, by Cumulativity, \mbox{$\cC(a , \neg b) = \cL$}
implies \mbox{$\cC(A , a , \neg b) = \cL$}, and, by the
Left Elimination rule for negation: \mbox{$b \in \cC(A , a)$}.
Property~\ref{forall} obviously implies~\ref{si}, that obviously
implies~\ref{mon}.
It is easy to see that property~\ref{mon} implies~\ref{logimpl}.
Let $m$ be any propositional model that satisfies $a$.
Let \cC\  be defined by \mbox{$\cC(A) = \overline{\{m\}}$},
the set of formulas satisfied by $m$,
if \mbox{$m \in \widehat{A}$} and \mbox{$\cC(A) = \cL$}
otherwise.
By assumption, \mbox{$b \in \cC(a)$}.
But \mbox{$\cC(a) = \overline{\{m\}}$} since 
\mbox{$m \in \widehat{\{a\}}$}, therefore
\mbox{$m \models b$}.

The only non-trivial part of the proof is that~\ref{logimpl}
implies~\ref{contra}.
Assume \mbox{$a \models b$} and \cC\  satisfies
Inclusion, Idempotence, Cautious
Monotonicity, Conditional Monotonicity and Threshold Monotonicity, Weak Compactness
and the 
Introduction-Elimination rules.
By Theorem~\ref{the:propcalc}, there is a set \cM, a satisfaction relation
$\models$ that behaves classically with respect to the connectives and
a definability-preserving choice function satisfying Contraction,
Coherence and Local Monotonicity such that 
\mbox{$\cC(a , \neg b) =$}
\mbox{$\overline{f(\widehat{\{a\}} \cap \widehat{\{\neg b\}})}$}.
But, by assumption 
\mbox{$\widehat{\{a\}} \cap \widehat{\{\neg b\}} = \emptyset$}.
By Contraction, then
\mbox{$\cC(a , \neg b) =$}
\mbox{$\overline{\emptyset} =$} \cL.
\QED
Theorem~\ref{the:prop} shows that the proof theory of the 
semantically-classical propositional connectives 
in a nonmonotonic setting is the same as in a monotonic setting.
The nonmonotonic setting is very rich, and it is tempting to consider,
there, connectives the semantics of which is not locally truth-functional:
the truth-value of \mbox{$\Box a$} or of \mbox{$a \succ b$}
in a model $m$ depending on the choice function $f$. 
This is left for future work.

It is customary to consider Introduction-Elimination rules
as definitions of the connectives.
Hacking~\cite[Section VII]{Hacking:What} discusses this idea
and proposes that, to be considered as bona fide definitions of the
connectives, the rules must be such that they
ensure that any legal logic on a small language may be
conservatively extended to a legal logic on the language extended
by closure under the connective.
We may ask whether any nonmonotonic operation 
\cC\  on \cL\ that satisfies Inclusion, Idempotence, Cautious Monotonicity,
Conditional Monotonicity and Threshold Monotonicity may be conservatively
extended to such an operation that satisfies
the Introduction-Elimination rules above: (\ref{eq:andrule}),
(\ref{eq:negleftintro}), (\ref{eq:negleftelim}), (\ref{eq:orleftintro}),
(\ref{eq:orrightintro}), (\ref{eq:imprightintro}) and 
{\ref{eq:impleftintro}).
From the discussion above, just before the definition of Weak Compactness,
it seems that Weak Compactness will be required.
The question of whether Weak Compactness is sufficient to ensure
a conservative extension is open.
The result will be proven under a stronger assumption: the set of
atomic propositions is finite (this essentially the Simplifying
Assumption of Section~\ref{subsec:simple}).
\begin{theorem}
\label{the:conser}
Let \cL\  be the propositional calculus on a {\em finite} set $P$ of atomic
propositions.
Let \cC\  be an operation on $P$ that satisfies
Inclusion, Idempotence,  Cautious Monotonicity,
Conditional Monotonicity and Threshold Monotonicity.
Then, there exists an operation \cC' on \cL\ that satisfies
Inclusion, Idempotence,  Cautious Monotonicity,
Conditional Monotonicity, Threshold Monotonicity and
the rules: (\ref{eq:andrule}),
(\ref{eq:negleftintro}), (\ref{eq:negleftelim}), (\ref{eq:orleftintro}),
(\ref{eq:orrightintro}), (\ref{eq:imprightintro}) and 
(\ref{eq:impleftintro}), such that, for any \mbox{$A \subseteq P$},
\mbox{$\cC(A) = P \cap \cC'(A)$}.
\end{theorem}
\proof
Since $P$ is finite, \cC\ is trivially weakly compact,
the assumptions of Theorem~\ref{the:propcalc} hold
and \cC\ is therefore generated by some \cM, $\models$ and $f$.
All subsets of $M$ are definable and $f$ is therefore defined on
all subsets of $M$.
We may extend $\models$ to the language \cL\  by using 
equations~(\ref{eq:and}), (\ref{eq:neg}), (\ref{eq:or}) and(~\ref{eq:imp}).
The choice function $f$, then defines an operation \cC' on \cL\
by (\ref{eq:fdef}).
Since $f$ satisfies Contraction, Coherence and Local Monotonicity,
the operation \cC' satisfies Inclusion, Idempotence,  Cautious Monotonicity,
Conditional Monotonicity and Threshold Monotonicity.
Since the models of $M$ satisfy equations~(\ref{eq:and}), (\ref{eq:neg}), 
(\ref{eq:or}) and(~\ref{eq:imp}), \cC\  satisfies the Introduction-Elimination
rules.
It is left to us to see that \mbox{$\cC(A) = P \cap \cC'(A)$},
for any \mbox{$A \subseteq P$}.
This follows straightforwardly from the fact that both $\cC(A)$
and $\cC'(A)$ are the set of formulas (the former of $P$, the latter
of \cL) satisfied by all models of the set \mbox{$f(\widehat{A})$}. 
\QED
\section{Comparison with previous work}
\label{sec:KLM}
Let us assume that \cL\ is a propositional language,
and that \cC\  satisfies Weak Compactness,
Inclusion, Idempotence, Cautious Monotonicity, Conditional Monotonicity,
Threshold Monotonicity and the Introduction-Elimination rules
for all the propositional connectives.
One may define a consequence operation on \cL\  by:
\mbox{$a \NI b$} iff \mbox{$b \in \cC(\{a\})$}.
\begin{theorem}
\label{the:pref1}
Under the assumptions above, the relation \NI\ is preferential
(see~\cite{KLMAI:89}).
\end{theorem}
\proof
The result follows easily from Theorem~\ref{the:propcalc}.
We shall treat only two of the six properties, the reader will
easily treat the other four properties.
Consider Left Logical Equivalence.
Assume that $a$ is logically equivalent to $a'$.
Then \mbox{$\widehat{\{a\}} = \widehat{\{a'\}}$}
and \mbox{$\cC(\{a\}) = \cC(\{a'\})$}.
Consider {\bf Or}.
Assume \mbox{$c \in \cC(\{a\})$} and \mbox{$c \in \cC(\{b\})$}.
Any \mbox{$x \in f(\widehat{\{a\}})$} satisfies $c$.
Any \mbox{$x \in f(\widehat{\{b\}})$} satisfies $c$.
Any $f$ satisfying Contraction and Coherence, satisfies
\mbox{$f(X \cup Y) \subseteq f(X) \cup f(Y)$}.
Therefore any 
\mbox{$x \in f(\widehat{\{a\}}) \cup f(\widehat{\{b\}})$} satisfies $c$
and \mbox{$c \in \cC(a \vee b)$}.
\QED
Does the converse hold, i.e., may any preferential relation be
obtained from such an operation \cC\ in such a way?
The answer is yes.
\begin{theorem}
\label{the:pref2}
Let \NI\  be any preferential relation.
There is an operation \cC\  satisfying all the assumptions
above such that \mbox{$a \NI b$} iff \mbox{$b \in \cC(\{a\})$}.
\end{theorem}
\proof
This follows from the construction of Theorem 14 of~\cite{FL:Studia}.
The theorem claims that any finitary preferential operation
(i.e., preferential relation) may be conservatively 
extended to an infinitary preferential operation \cC.
The definition of \cC\ is the following:
\mbox{$b \in \cC(A)$} iff there exists some formula $a$ such that
\mbox{$A \models a$} ($\models$ is logical implication of
propositional calculus) enjoying the following property:
for any $a'$ such that \mbox{$A \models a'$} and \mbox{$a' \models a$},
one has \mbox{$a' \NI b$}.

The reader may check, with no need to use the theorem of~\cite{FL:Studia}
and relatively easily, that \cC\  satisfies all the properties
requested.
For example, \cC\ is weakly compact.
Assume \mbox{$\cC(A) = \cL$}. Then \mbox{${\bf false} \in \cC(A)$}
and there is some $a$ such that \mbox{$A \models a$} and
\mbox{$a \NI {\bf false}$}.
But there is a finite subset $B$ of $A$ such that 
\mbox{$B \models a$} and \mbox{$B \NI {\bf false}$}.
Therefore \mbox{${\bf false} \in \cC(B)$}. 
\QED 
\section{Rational Monotonicity}
\label{sec:rat}
In this section, an important sub-family of nonmonotonic operations
will be described. It corresponds, in the present setting,
to the rational relations of~\cite{LMAI:92}.
The qualitative measures of Section~\ref{subsec:qualmeas}
were partial orders.
Measures (probability measures, for example) provide
orders that obey an additional modularity property:
\begin{equation}
\label{eq:modular}
X > Y , \: X \not > Z \Rightarrow Z > Y. 
\end{equation}
This seems a very natural property to require
of a qualitative measure.
Assume for example that there is a function 
\mbox{$m : 2^{\cM} \longrightarrow {\bf R}$}, the set of
real numbers such that \mbox{$X > Y$} iff \mbox{$m(X) > m(Y)$},
then, $>$ is modular.

What is the property of choice functions that correspond
to the modularity of a qualitative measure?
\[
{\bf Arrow} \ \ X \subseteq Y , \: X \cap f(Y) \neq \emptyset
\: \Rightarrow \: f(X) \subseteq f(Y)
\]
This property is in fact only one half of the property studied
by K. Arrow in~\cite{Arrow:59}, assuming Contraction.
The other half is Coherence.
Note indeed that Arrow, Coherence and Contraction imply
\begin{equation} \label{eq:Arroworiginal}
X \subseteq Y , \: X \cap f(Y) \neq \emptyset
\: \Rightarrow \: f(X) = X \cap f(Y),
\end{equation}
the property originally considered by K. Arrow.
Indeed, by Coherence \mbox{$X \cap f(Y) \subseteq$} \mbox{$f(X)$};
by Arrow, \mbox{$f(X) \subseteq$} \mbox{$f(Y)$};
by contraction \mbox{$f(X) \subseteq$} \mbox{$X$}.

The intuitive justification for Arrow is some kind of 
laziness principle for the choice function:
if \mbox{$X \subseteq Y$} and we have already a list of the best elements
of $Y$, we shall take for the best elements of $X$ exactly
those best elements of $Y$ that happen to be in $X$, at least whenever 
this new list is not empty.
A remark will now show that there is a natural family of choice functions
that satisfy Arrow.
Suppose the elements of \cM\ are ranked: our real estate agent,
for example, gives a grade to every available apartment and, when
asked about apartments in some area, delivers the list of all available
apartments in this area {\em that have the highest ranking}.
If the highest ranking in Paris is 10 and one of those apartments
graded 10 happens to be on the left bank, then all best apartments
on the left bank have grade 10 and are therefore part of the list
of best apartments in Paris.
We shall now show that Modularity and Arrow are indeed exact counterparts.
\begin{theorem}
\label{the:ModArrow}
If $>$ satisfies 
properties~\ref{eq:inc} and~\ref{eq:modular}, 
then, the choice function $f$ defined by taking for
$f(X)$ the set of heavy elements (see Definition~\ref{def:heavy}) of $X$
satisfies Arrow.
If $f$ is a choice function that satisfies Contraction, Coherence, 
Local Monotonicity and Arrow,
then, the relation $>$ defined by:
\mbox{$X > Y$} iff \mbox{$f(X) \neq \emptyset$} and 
\mbox{$Y \cap f(X \cup Y) = \emptyset$} satisfies 
property~\ref{eq:modular}.
\end{theorem}
\proof
Assume $>$ satisfies 
properties~\ref{eq:inc} and~\ref{eq:modular}, 
and that \mbox{$X \subseteq Y$} and \mbox{$x \in X$} such that
\mbox{$Y \not > \{ x \}$}.
Let $x'$ be any heavy element of $X$: \mbox{$X \not > \{ x' \}$}.
If $x'$ was not heavy in $Y$, we would have \mbox{$Y > \{ x' \}$}
and, by~\ref{eq:modular}, \mbox{$Y > X$} and, by~\ref{eq:inc},
\mbox{$Y > \{ x \}$}, a contradiction.
Therefore any heavy element of $X$ is a heavy element of $Y$.

Assume now that $f$ satisfies Contraction, Coherence, Local
Monotonicity and Arrow and that \mbox{$X > Y$}.
We shall show that either \mbox{$X > Z$} or \mbox{$Z > Y$}.
We know that \mbox{$f(X) \neq \emptyset$} and
\mbox{$Y \cap f(X \cup Y) = \emptyset$}. 
We distinguish two cases.
Assume, first, that \mbox{$(Y \cup Z) \cap f(X \cup Y \cup Z) \neq \emptyset$}.
By Arrow, then, \mbox{$f(Y \cup Z) \subseteq$}
\mbox{$f(X \cup Y \cup Z)$}.
But, by Contraction and then Coherence
\[
f(X \cup Y \cup Z) = f(X \cup Y \cup (Z - Y)) =
\]
\[
(X \cup Y) \cap f(X \cup Y \cup (Z - Y)) 
\cup (Z - Y) \cap f(X \cup Y \cup (Z - Y)) \subseteq
f(X \cup Y) \cup (Z - Y).
\]
Therefore
\mbox{$Y \cap f(Y \cup Z) = \emptyset$}.
If \mbox{$f(Z) \neq \emptyset$}, then, we have \mbox{$Z > Y$}.
If \mbox{$f(Z) = \emptyset$}, since, by Coherence,
\mbox{$Z \cap f(X \cup Z) \subseteq f(Z)$},
we have \mbox{$X > Z$}.

We are left with the case 
\mbox{$(Y \cup Z) \cap f(X \cup Y \cup Z) = \emptyset$}.
By Contraction, \mbox{$f(X \cup Y \cup Z) \subseteq$}
\mbox{$X \subseteq$}
\mbox{$X \cup Z \subseteq$}
\mbox{$X \cup Y \cup Z$}.
By Local Monotonicity, then, 
\mbox{$f(X \cup Z) \subseteq f(X \cup Y \cup Z)$}.
Therefore \mbox{$Z \cap f(X \cup Z) = \emptyset$}
and we conclude that \mbox{$X > Z$}.
\QED
Having proved the equivalence of Modularity and Arrow,
we shall prove equivalence Modularity and the property of
Rational Monotonicity for the operation \cC. 
\begin{theorem}
\label{the:rat}
Suppose we are given a language \cL \ and a function 
\mbox{$\cC : 2^{\cL} \longrightarrow 2^{\cL}$}.
Then, the following two conditions are equivalent:
\begin{enumerate}
\item \label{Carrow}
\cC\  satisfies Inclusion, Idempotence, Cautious Monotonicity, Conditional Monotonicity,
Threshold Monotonicity and
\[
{\bf Rational \ Monotonicity} \ \  
\cC(\cC(A) , B) \neq \cL \: \Rightarrow \: 
\cC(A) \subseteq \cC(A , B),
\]
\item \label{farrow}
there exists a set \cM \ (of models), a satisfaction relation
\mbox{$\models \: \subseteq \cM \times \cL$} and a definability-preserving
choice function
\mbox{$f : \cM \longrightarrow \cM$} 
satisfying Contraction, Coherence, Local Monotonicity and Arrow such that
\mbox{$\cC(A) =$} \mbox{$\overline{f(\widehat{A}})$}. 
\end{enumerate}
\end{theorem}

Rational Monotonicity, introduced in~\cite{KLMAI:89}, has been studied
at length in~\cite{LMAI:92,FL:Studia}.
Its intuitive justification is that, when the new information,
contained in $B$, is consistent with what had been concluded
from $A$, $\cC(A)$, the new information will not force us to
retract any previous conclusions, it may only add new conclusions.
This is also a laziness principle.
\proof
We shall describe only those parts of the proof that differ from that
of Theorem~\ref{the:main}.
For the soundness part: \ref{farrow} implies~\ref{Carrow}, 
define \mbox{$\cC(A)$} as \mbox{$\overline{f(\widehat{A})}$} and
assume Arrow. We must derive Rational Monotonicity.

Assume 
\mbox{$\cC(\cC(A) , B) \neq \cL$}.
On one hand we have \mbox{${\rm Mod}(A \cup B) \subseteq \widehat{A}$}
and on the other hand we have
\mbox{$\overline{f({\rm Mod}(\cC(A) \cup B))} \neq \cL$}.
Therefore,   
\mbox{$f({\rm Mod}(\cC(A) \cup B)) \neq \emptyset$}
and, by Contraction,
\mbox{${\rm Mod}(\cC(A) \cup B) \neq \emptyset$}, in other terms,
\mbox{$\widehat{\cC(A)} \cap \widehat{B} \neq \emptyset$}.
Since $f$ preserves definability:
\mbox{$\widehat{\cC(A)} =$}
\mbox{$f(\widehat{A})$}.
We see that
\mbox{$f(\widehat{A}) \cap \widehat{B} \neq \emptyset$}
and, by Contraction,
\mbox{$f(\widehat{A}) \cap {\rm Mod}(A \cup B) \neq \emptyset$}.
By Arrow, \mbox{$f({\rm Mod}(A \cup B)) \subseteq$}
\mbox{$f(\widehat{A})$}.
Therefore \mbox{$\cC(A) \subseteq$}
\mbox{$\cC(A , B)$}.

For the completeness part, we define \Cn\ as in the proof
of Theorem~\ref{the:main}. The properties of \Cn\ proved there
hold true.
We must define \cM\  slightly more carefully.
We take for elements of \cM\  only those sets \mbox{$T \neq \cL$}
such that \mbox{$T = \Cn(T)$}.
The satisfaction relation is defined as previously:
\mbox{$T \models a$} iff \mbox{$a \in T$}.
We see that, by construction, no element of \cM\  satisfies all formulas
of \cL\ and therefore \mbox{$\overline{X} = \cL$} implies 
\mbox{$X = \emptyset$}.
We define $f$ as previously. The proof that Equation~\ref{eq:m}
holds is unchanged.
Assuming Rational Monotonicity, we must show that Arrow holds.
Assume
\mbox{$X \subseteq$}
\mbox{$Y$} and
\mbox{$X \cap f(Y) \neq \emptyset$}.
We have \mbox{$\widehat{\overline{X}} \subseteq$}
\mbox{$\widehat{\overline{Y}}$}
and \mbox{$X \cap f(Y) =$}
\mbox{$X \cap Y \cap \widehat{\cC(\overline{Y})} =$}
\mbox{$X \cap \widehat{\cC(\overline{Y})} \neq$} 
\mbox{$\emptyset$}.
Therefore 
\mbox{$\widehat{\overline{X}} \cap \widehat{\cC(\overline{Y})} \neq$} 
\mbox{$\emptyset$}.
Let us define \mbox{$A = \overline{X}$} and
\mbox{$B = \overline{B}$}.
On one hand, we have 
\mbox{$\widehat{A} \subseteq$} \mbox{$\widehat{B}$}
and therefore \mbox{$\Cn(B) \subseteq \Cn(A)$}
and on the other hand we have
\mbox{$\widehat{A} \cap \widehat{\cC(B)} \neq$} 
\mbox{$\emptyset$}.
Therefore 
\mbox{${\rm Mod}(A \cup \cC(B)) \neq \emptyset$}.
By the remark made at the start of this proof, 
when paying attention to exclude
\cL\  from the set \cM, we see that
\mbox{$\overline{{\rm Mod}(A \cup \cC(B))} \neq \cL$}.
Therefore 
\mbox{$\cC(A , \cC(B)) \neq \cL$}.
It is easy to show that
\mbox{$\cC(A , \cC(B)) = \cC(\Cn(A) , \cC(\Cn(B)))$}.
We may therefore use Rational Monotonicity to conclude that we have
\mbox{$\cC(B) \subseteq \cC(A)$}, i.e.,
\mbox{$\cC(\overline{Y}) \subseteq$} \mbox{$\cC(\overline{X})$} and
\mbox{$\widehat{\cC(\overline{X})} \subseteq$}
\mbox{$\widehat{\cC(\overline{Y})}$}.
Therefore
\mbox{$X \cap \widehat{\cC(\overline{X})} \subseteq$}
\mbox{$Y \cap \widehat{\cC(\overline{Y})}$}, i.e.,
\mbox{$f(X) \subseteq$} \mbox{$f(Y)$}.
\QED
We have shown that the additional property of Rational Monotonicity, 
studied in the literature, corresponds exactly to an additional
property of the choice function $f$.
The family of nonmonotonic operations satisfying Rational Monotonicity
has different closure properties than the larger family studied
in the preceding sections. 
In particular, it is not closed under intersection.
The reader will easily find a counter example.
One such example is provided in~\cite{Mak:Handbook}.
\section{Conclusion}
\label{sec:conc}
We have described two quite different but equivalent semantic frameworks: 
choice functions and qualitative measures, that provide an ontology for
nonmonotonic deduction.
Choice functions have been studied by researchers in Social Choice
for their {\em rationalizability} properties, i.e., by what kind of
aggregation mechanism can they arise from individual preferences?
The equivalence we have shown with qualitative measures may be of interest
to those researchers.
The family of nonmonotonic operations defined in Section~\ref{sec:nonmonded}
is precisely the family defined by choice functions or by qualitative
measures, it is a natural generalization of Tarski's monotonic deductive
operations.
The operations of this family are closed under intersection.
The classical connectives may be defined elegantly for this family
of operations, by properties that are weaker than those generally
considered in the monotonic case.
Only a very mild compactness assumption is needed.
The sentential connectives may be {\em defined}
by Introduction-Elimination rules. The connectives defined have a
classical semantics.
A further property of choice functions, considered by K. Arrow~\cite{Arrow:59},
is shown to be equivalent to the modularity (i.e., negative-transitivity)
of the qualitative measure and the operations defined are characterized by
the additional property of Rational Monotonicity of~\cite{KLMAI:89,LMAI:92}.

A large number of questions and alleys for future research are left open
by this work.
Let us mention a few, roughly from the small and technical to the 
vast and philosophical.

The equivalence of Qualitative Measures and Choice functions semantics
has been shown only under the Simplifying Assumption.
A more general equivalence requires the introduction of a family
of definable sets in the framework of Qualitative Measures and
probably the introduction of some counterpart to Definability Preservation.

Theorem~\ref{the:conser} has been proved only under the assumption
that $P$ is finite. Does it hold without this restriction, and if not,
does it hold without this restriction if one assumes \cC\ to be
weakly compact?

The framework of Choice Functions begs the semantic 
definition of non-classical, non-truth functional
connectives.
The study of such connectives (unary or, more probably, binary) 
in nonmonotonic logics 
seems particularly exciting. The {\em preferred} interpretation
of the choice function suggests a link with deontic logics.

This work sheds new light on properties studied by researchers in
Social Choice. In particular new insights on the case of an infinite set
of outcomes have been presented. Are they relevant to the Social
Choice community?

In\cite{LMAI:92}, a positive answer was given to a question that can now be
seen as equivalent to: given a choice function that satisfies
Contraction, Coherence and Local Monotonicity, is there a canonical
way to restrict this choice function in a way that ensures the 
Arrow property? This canonical construction, rational closure,
offers a way to aggregate individual preferences into collective
preferences that satisfy the Arrow property. This aggregation
method does not satisfy Independence from Irrelevant Alternatives.
Is it of any interest for Social Choice?

This work uses Tarski's framework. Most proof-theoretic studies
use Gentzen's framework. The translation of the results of this paper
to the language of Gentzen's sequents may be illuminating.
It seems one will have to consider sequents whose sides may be infinite
sets of formulas. 

\section{Acknowledgments}
A number of people provided me with suggestions that improved
the presentation of this paper, in particular, Shai Berger,
Tom Costello, David Israel,
Michael Freund, Karl Schlechta and an anonymous referee.
Very special thanks are due to David Makinson for his remarks, both conceptual
and technical and to the students of CS67999 (5758 edition) who provided
the motivation for this work and very useful comments.
\bibliographystyle{plain}

\end{document}